%% file: arxiv.tex
\documentclass{article}

\let\vimodaddcontentsline\addcontentsline
\usepackage{iclr2027_conference,times}
\let\addcontentsline\vimodaddcontentsline
\input{styles/math_commands.tex}

\usepackage{amsmath}
\usepackage{amssymb}
\usepackage{booktabs}
\input{styles/table_style.tex}
\usepackage{capt-of}
\usepackage{fvextra}
\usepackage[most]{tcolorbox}
\usepackage{graphicx}
\usepackage{needspace}
\usepackage{wrapfig}
\usepackage{fix-cm}
\usepackage{hyperref}
\usepackage{microtype}
\usepackage{url}
\usepackage{algorithm}
\usepackage{algorithmic}
\usepackage{enumitem}

\let\vimodoriginalttdefault\ttdefault
\usepackage[narrow]{inconsolata}
\let\ttdefault\vimodoriginalttdefault
\DeclareRobustCommand{\codeparam}[1]{{\fontencoding{T1}\fontfamily{zi4}\selectfont #1}}
\DeclareRobustCommand{\modelname}[1]{\codeparam{#1}}
\hypersetup{
  hypertexnames=false,
  colorlinks=false,
  pdfborder={0 0 1},
  citebordercolor={0 1 0},
  linkbordercolor={1 0 0},
  urlbordercolor={0 1 1}
}
\newif\ifvimodappendixcontents
\let\vimodlinkedaddcontentsline\addcontentsline
\renewcommand{\addcontentsline}[3]{%
  \ifvimodappendixcontents\vimodlinkedaddcontentsline{#1}{#2}{#3}\fi}

\makeatletter
\let\vimod@makecaption\@makecaption
\long\def\@makecaption#1#2{{\small\vimod@makecaption{#1}{#2}}}
\let\vimod@floatc@ruled\floatc@ruled
\long\def\floatc@ruled#1#2{{\small\vimod@floatc@ruled{#1}{#2}}}
\makeatother

\newcommand{\figcaptionsep}{\vspace{-1em}}
\graphicspath{{figures/}}

\title{Look Back, Think Ahead: Visual Memory on Demand for Efficient Multimodal Reasoning}

\author{
    \textbf{Yicheng Xue}\textsuperscript{1,2},
    \textbf{Han Wu}\textsuperscript{3},
    \textbf{Jufeng Yang}\textsuperscript{4},
    \textbf{Minjing Dong}\textsuperscript{1},
    \textbf{Xinghao Chen}\textsuperscript{5},\\
    \textbf{Hanting Chen}\textsuperscript{5},
    \textbf{Jianyuan Guo}\textsuperscript{1*}
    \\[0.5em]
    {\normalsize
        \textsuperscript{1}City University of Hong Kong,
        \textsuperscript{2}Zhejiang University,
        \textsuperscript{3}Peking University,
    }\\
    {\normalsize
        \textsuperscript{4}Nankai University,
        \textsuperscript{5}Huawei Technologies
    }
    \\[0.2em]
    {\normalsize\texttt{yichengxue@zju.edu.cn, jianyguo@cityu.edu.hk}}
}

\iclrfinalcopy

\input{styles/ablation_style}

\input{tables/ablation_panels}
\begin{document}

\maketitle
\begingroup
    \renewcommand{\thefootnote}{*}
    \footnotetext{Corresponding author.}
\endgroup

\fancyhead{}
\renewcommand{\headrulewidth}{0pt}

\begin{abstract}
\input{sections/abstract}
\end{abstract}

\input{sections/introduction}
\input{sections/related_work}
\input{sections/method}
\input{sections/experiments}
\input{sections/limitations}
\input{sections/conclusion}

\bibliography{references}
\bibliographystyle{iclr2027_conference}

\appendix
\clearpage
\vimodappendixcontentstrue
\begin{center}
  {\LARGE\bfseries Supplementary Material\par}
\end{center}
\vspace{1em}
\setcounter{tocdepth}{2}
\tableofcontents
\input{appendices/additional_method}
\input{appendices/training_details}

\input{appendices/experimental_details}
\input{appendices/qualitative_analysis}
\input{appendices/teacher_prompts}

\end{document}

%% file: styles/math_commands.tex
\usepackage{amsmath,amsfonts,bm}

\def\eqref#1{equation~\ref{#1}}

\def\1{\bm{1}}

\DeclareMathAlphabet{\mathsfit}{\encodingdefault}{\sfdefault}{m}{sl}
\SetMathAlphabet{\mathsfit}{bold}{\encodingdefault}{\sfdefault}{bx}{n}



%% file: styles/table_style.tex
\usepackage{array}
\usepackage{colortbl}
\usepackage{multirow}
\usepackage{tabularx}
\usepackage{environ}
\newcolumntype{E}{>{\raggedleft\arraybackslash}X}
\definecolor{tablegroupgray}{gray}{0.92}

\newcolumntype{C}{>{\centering\arraybackslash}X}
\definecolor{expheader}{gray}{0.875}
\definecolor{expcost}{gray}{0.875}
\definecolor{expours}{rgb}{1,0.85,0.85}
\definecolor{expband}{gray}{0.945}
\NewEnviron{experimenttable}{%
  \par\vspace{10.135pt}%
  \begingroup
  \centering
  \normalsize
  \setlength{\tabcolsep}{3.5pt}%
  \renewcommand{\arraystretch}{1}%
  \setlength{\arrayrulewidth}{0.4pt}%
  \setlength{\heavyrulewidth}{0.8pt}%
  \setlength{\lightrulewidth}{0.5pt}%
  \setlength{\cmidrulewidth}{0.4pt}%
  \setlength{\aboverulesep}{1.7222pt}%
  \setlength{\belowrulesep}{2.7986pt}%
  \scalebox{0.6236}{\BODY}%
  \par\endgroup
}

\newenvironment{mainresultstable}{%
  \par\smallskip\begingroup\centering
  \fontsize{10}{11}\selectfont
\definecolor{header}{gray}{0.875}
\definecolor{budgetband}{gray}{0.92}
\definecolor{budgetaccent}{HTML}{D98200}
\definecolor{best}{rgb}{1,0.85,0.85}
\definecolor{second}{rgb}{0.85,0.85,1}
\definecolor{ratio}{gray}{0.35}
\setlength{\tabcolsep}{2.3pt}
\setlength{\heavyrulewidth}{0.8pt}
\setlength{\lightrulewidth}{0.5pt}
\setlength{\arrayrulewidth}{0.4pt}
\setlength{\aboverulesep}{1.7222pt}
\setlength{\belowrulesep}{2.7986pt}
\renewcommand{\arraystretch}{1}
}{\par\endgroup}

%% file: styles/ablation_style.tex
\newenvironment{ablationtable}[1][4.8pt]{%
  \par\begingroup\centering\normalfont\fontsize{8}{8.8}\selectfont
  \setlength{\tabcolsep}{#1}
  \renewcommand{\arraystretch}{1}
  \setlength{\heavyrulewidth}{0.48pt}
  \setlength{\lightrulewidth}{0.3pt}
  \setlength{\cmidrulewidth}{0.18pt}
  \setlength{\aboverulesep}{1.376pt}
  \setlength{\belowrulesep}{2.24pt}
  \setlength{\abovetopsep}{0pt}
  \setlength{\belowbottomsep}{0pt}
  \setlength{\doublerulesep}{1.6pt}
}{%
  \par\endgroup
}
\newcommand{\abudget}[1]{#1\%}

%% file: sections/abstract.tex
Processing long visual token sequences from high-resolution images makes multi-step reasoning computationally expensive for multimodal Large Language Models (MLLMs). Existing one-shot pruning and aggregation methods compress visual tokens into a fixed context before decoding. However, visual evidence needs can shift as reasoning unfolds, making it difficult for a fixed compressed context to retain all the details needed across stages. To address this challenge, we propose ViMoD, a lightweight framework that maintains a compact visual context while preserving access to original fine-grained evidence as reasoning needs evolve. Deformable Aggregation of Region-wise Tokens (DART) learns content-adaptive groups and aggregation capacities, constructing compact Coarse representations linked to recoverable original Fine tokens. Temporal Routing for Adaptive Contextual Evidence (TRACE) integrates decoding history to anticipate upcoming evidence needs and select, retain, or replace active Fine-token groups. Selected Fine tokens augment the persistent Coarse context in the frozen backbone, enabling stage-specific evidence access without continuously attending to all visual tokens. On Qwen3-VL-4B, ViMoD outperforms all evaluated baselines on all eight reasoning benchmarks at a \textbf{20\%} target visual token budget, improving the mean normalized score by \textbf{39.0\%} over the strongest evaluated one-shot baseline. These gains are achieved with only \textbf{0.0546\%} additional trainable parameters relative to the frozen backbone.

%% file: sections/introduction.tex
\section{Introduction}
\label{sec:introduction}

Multimodal Large Language Models (MLLMs) \citep{bai2025qwen3,wang2025internvl3,team2025kimi,hong2025glm} support diverse visual understanding and reasoning tasks. However, processing long visual token sequences incurs substantial inference costs, particularly for high-resolution inputs. Visual token compression offers a practical way to reduce these costs.

Existing methods, including FastV \citep{chen2024image}, VisionZip \citep{yang2025visionzip}, and DivPrune \citep{alvar2025divprune}, reduce this cost by pruning or merging visual tokens. Many compress the visual context early in inference and keep the resulting compact representation fixed throughout decoding, a strategy we refer to as \textbf{one-shot visual compression}. 
Although this strategy reduces computation per generated token, it raises two questions for multi-step visual reasoning: \emph{Can a fixed compressed context efficiently support changing visual evidence needs?} \emph{Do per-token savings reduce total decoding cost over a complete response?}

\begin{figure}[t]
  \centering
  \includegraphics[width=\linewidth]{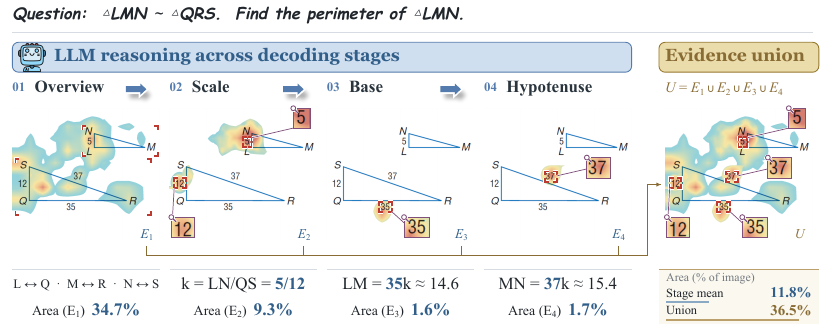}
  \vspace{-24pt}
  \caption{\small{\textbf{Visual contribution changes across different reasoning stages.} The model first forms an overview of the triangle diagram, determines the scale factor from 5 and 12, and reads the known base and hypotenuse, 35 and 37. Heatmaps show gradient-based visual token contribution at selected decoding stages; warmer colors indicate higher contribution relative to other tokens in the same stage. The right panel combines the corresponding regions into $U=E_1\cup E_2\cup E_3\cup E_4$. These regions occupy 11.8\% of the image on average per stage, whereas their union occupies 36.5\%, counting overlapping areas once.}}
  \label{fig:dynamic-demand}
\end{figure}

\begin{figure}[t]
  \centering
  \begin{minipage}[t]{0.405\linewidth}
    \centering
    {\small\textbf{(a) Answer retention}}\par
    \includegraphics[
      trim=2.5bp 1.5bp 0bp 2.5bp,
      clip,width=\linewidth
    ]{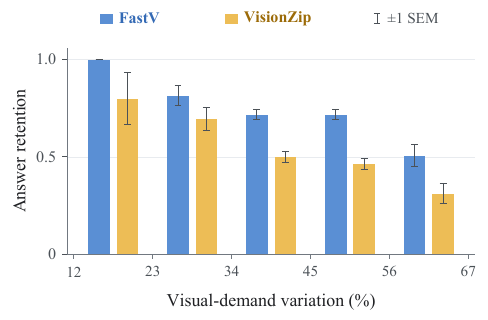}
  \end{minipage}
  \hspace{0.02\linewidth}
  \begin{minipage}[t]{0.405\linewidth}
    \centering
    {\small\textbf{(b) Evidence access}}\par
    \includegraphics[
      trim=2.5bp 1.5bp 0bp 2.5bp,
      clip,width=\linewidth
    ]{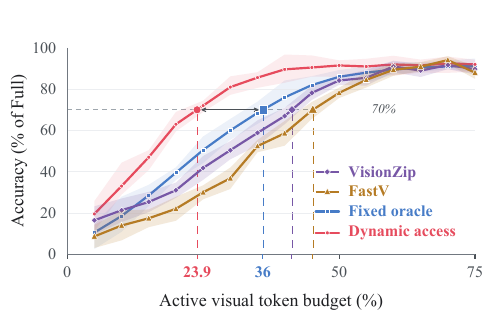}
  \end{minipage}

  \caption{\textbf{Visual demand and evidence access.}
  (a) On MathVerse at 50\% token retention, the fraction of
  baseline-correct examples remaining correct after compression,
  grouped by visual-demand variation; error bars show SEM.
  (b) On MMMU-Pro, the same \modelname{Qwen3.8}-annotated regions
  are accessed stage by stage (Dynamic access) or as their union
  (Fixed oracle), alongside VisionZip and FastV.
  Accuracy is normalized to Full; bands show standard deviation.}
  \label{fig:evidence-access}
  \label{fig:answer-retention}
  \label{fig:access-tradeoff}
\end{figure}

Visual evidence needs can shift across successive stages of reasoning. In the triangle-perimeter example in Fig.~\ref{fig:dynamic-demand}, regions with high gradient-based visual contribution change across successive portions of the response. We quantify this variation using the average turnover of highly attributed visual tokens between adjacent decoding windows as a proxy for changes in visual demand (Appendix~\ref{app:diagnostic-settings}). Larger shifts are associated with lower answer retention when 50\% of visual tokens are retained (Fig.~\ref{fig:answer-retention}(a)). 
This association suggests that a fixed compressed representation may fail to preserve evidence needed at different reasoning stages.

Even with advance knowledge of the required regions, a fixed context can require a larger visual token budget than stage-specific access. To examine this, we compare Dynamic access, which provides annotated regions at their corresponding reasoning stages, with Fixed oracle, which receives the union of all stage-specific annotated regions at the start of generation. 
At 70\% of Full's accuracy, the interpolated visual token budgets are approximately 23.9\% for Dynamic access and 36.0\% for Fixed oracle (Fig.~\ref{fig:evidence-access}(b)).
This comparison suggests that efficient visual compression depends not only on which evidence is retained, but also on how that evidence is made available across reasoning stages.

\begin{figure}[bt]
  \centering
  \includegraphics[trim=5bp 0bp 3bp 2bp,clip,width=0.90\linewidth]{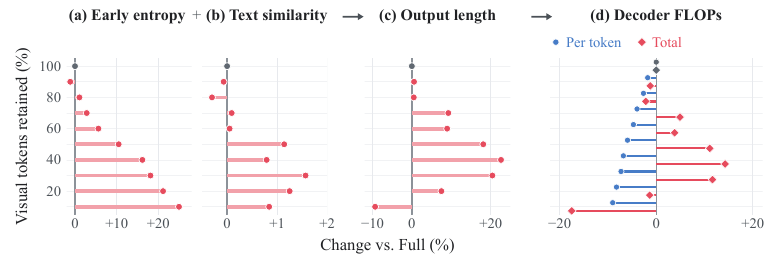}
  \figcaptionsep
  \caption{\small{\textbf{From per-token savings to response-level cost.} FastV on MathVerse; horizontal axes show percentage changes relative to the uncompressed baseline. \textbf{(a)} Early next-token entropy reflects predictive uncertainty. \textbf{(b)} Neighboring-window embedding similarity measures the similarity of successive content. \textbf{(c)} Output length counts generated tokens. \textbf{(d)} Estimated decoder FLOPs compare amortized cost per generated token (circles) with total response cost (diamonds). At several retention ratios, longer responses outweigh per-token savings, increasing total decoder computation.}}
  \label{fig:reasoning-cost}
\end{figure}

Per-token savings also do not necessarily reduce total decoding cost. For FastV on MathVerse, longer responses at several retention ratios offset the reduction in FLOPs per generated token, increasing estimated total decoder FLOPs relative to the uncompressed baseline (Fig.~\ref{fig:reasoning-cost}). Compression efficiency must therefore be evaluated over complete responses.

Together, these findings motivate a framework that adapts visual evidence access to evolving reasoning needs while reducing computation over complete responses. Recoverable visual memory enables a compact active context without permanently discarding fine-grained evidence. Existing recovery methods such as DSTP \citep{kim2026and} rely on decoding-time attention changes. This signal reflects the model's current interaction with the visual context, whereas recalled evidence may remain active over subsequent steps. We therefore seek to learn both a compact, recoverable visual memory and a recall policy that uses decoding history to anticipate upcoming evidence needs.

To this end, we propose \textbf{ViMoD: Visual Memory on Demand}, coupling learned visual memory with adaptive access to fine-grained visual tokens (Fine) and targeting a low average active visual token budget during decoding. \textbf{Deformable Aggregation of Region-wise Tokens (DART)} learns region-wise capacities and deformable groups, linking compact Coarse representations to recoverable Fine-token groups through a \textbf{Coarse-to-Fine (C2F)} map. \textbf{Temporal Routing for Adaptive Contextual\nopagebreak[4] Evidence (TRACE)} uses a state-space model to integrate decoding history, anticipate upcoming evidence needs, and select, retain, or replace active Fine-token groups. Joint-KV combines selected Fine tokens with persistent Coarse context in the frozen backbone's attention.

ViMoD achieves state-of-the-art performance in visual reasoning under limited visual token budgets (Tab.~\ref{tab:main}). At a 20\% target budget with 9:1 memory, its average normalized score exceeds DSTP$^\dagger$ on the same DART memory by 10.1\% and DivPrune, the strongest one-shot baseline, by 39.0\%. In our MMMU-Pro budget sweep, ViMoD reduces total inference time relative to Full at all evaluated target budgets from 20\% to 90\% (Fig.~\ref{fig:efficiency-tradeoff}).

%% file: sections/related_work.tex
\Needspace{8\baselineskip}
\section{Related Work}
\label{sec:related-work}

\noindent\textbf{Visual token compression and recovery.}
One-shot compression \citep{chen2024image,yang2025visionzip,song2026ktv,lu2026rora,alvar2025divprune,zhang2026d2pruner} prunes or merges visual tokens early, keeping the compressed context fixed throughout decoding. Recovery methods \citep{kim2026and} reintroduce tokens using attention changes, although visual attention sinks and head-specific behavior \citep{kang2025see,kang2025your} complicate attention-based evidence selection. ViMoD couples learned compression with a dynamic Fine-access policy supervised by upcoming reasoning needs, maintaining a compact, persistent Coarse context.

\noindent\textbf{Efficient text compression and deformable sampling.}
Gist-based LLM methods compress text into compact representations \mbox{\citep{mu2023learning,deng2026unigist}} and selectively recover original tokens \citep{mao2026simplified}. ViMoD adapts this combination of persistent compressed context and on-demand recovery to images, accounting for spatial structure and region-dependent granularity. Inspired by content-dependent sampling in deformable convolutions \citep{dai2017deformable}, DART learns region-wise capacities and deformable groups shared by Coarse aggregation and C2F-guided Fine recall.

\noindent\textbf{Interleaved visual reasoning.}
During generation, the tool-based DeepEyes~\citep{zheng2026deepeyes} crops and encodes local regions for additional high-resolution observations, while explicit interaction formats \citep{hu2026tvi} train the backbone to emit specialized tokens controlling transitions between textual reasoning and visual access. ViMoD instead learns access decisions through lightweight auxiliary modules and integrates selected evidence through Joint-KV, keeping the backbone frozen.

%% file: sections/method.tex
\section{Method}
\label{sec:method}

\subsection{Visual Memory and Evidence Access}
\begin{figure}[t]
  \centering
  \includegraphics[trim=12bp 12bp 25bp 48bp,clip,width=\linewidth]{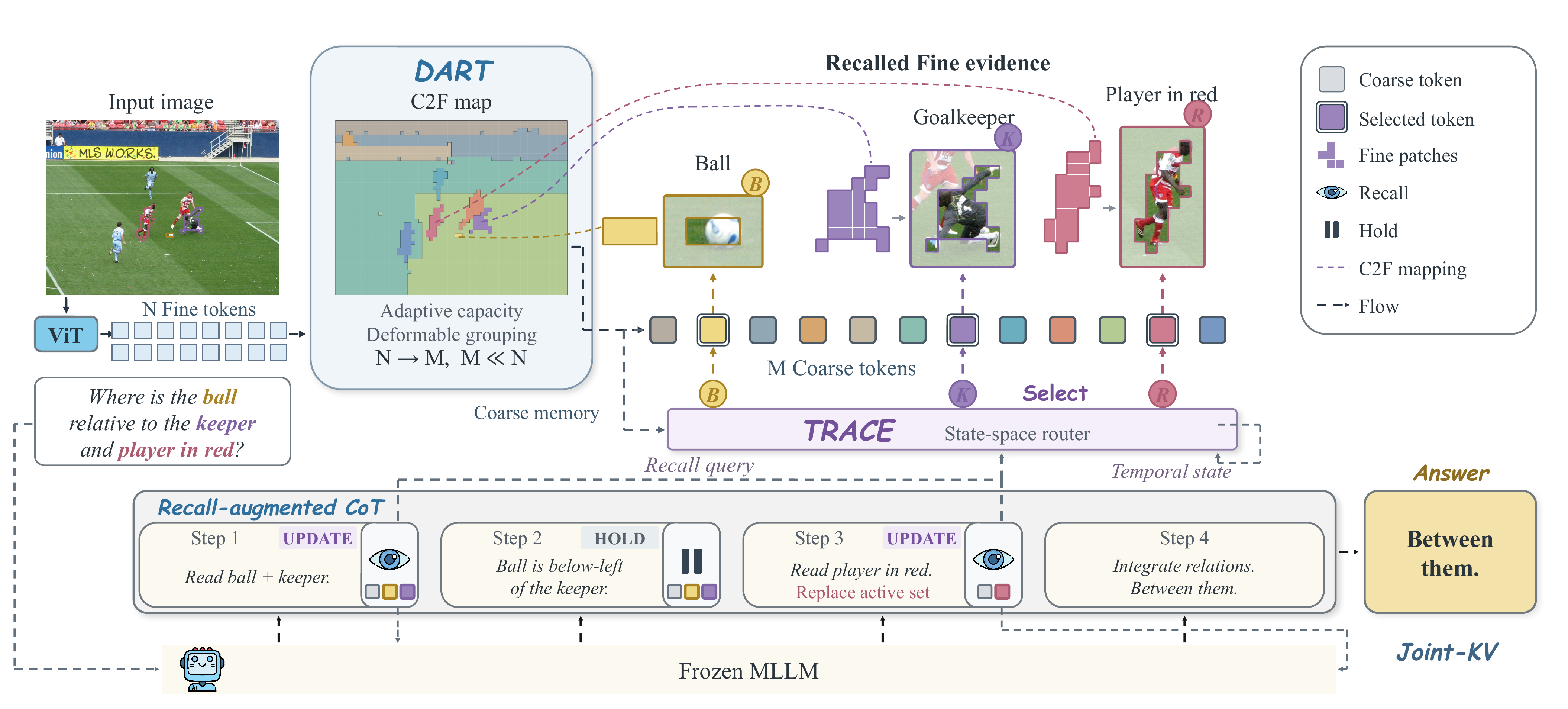}
  \figcaptionsep
  \caption{\textbf{Overview of ViMoD.} DART constructs persistent Coarse context and a C2F mapping to the original Fine-token groups. TRACE proposes evidence groups and retains or replaces the active set. Joint-KV exposes the selected Fine evidence alongside Coarse context and text history.}
  \label{fig:overview}
\end{figure}

Given an image $I$ and prompt $x$, ViMoD performs prefill using a multimodal backbone with frozen parameters $\theta_{\mathrm{Frozen}}$ and retains the original visual and text key--value (KV) banks (Fig.~\ref{fig:overview}).

Let $[N]=\{1,\ldots,N\}$ index the Fine embeddings $\mathbf X^{\mathrm F}\in\mathbb R^{N\times d}$. DART produces $M\ll N$ Coarse descriptors $\mathbf X^{\mathrm C}\in\mathbb R^{M\times d}$ for routing, corresponding layer-wise KV entries for attention, and a C2F map $\Pi(j)\subseteq[N]$ linking each group to its Fine members. At routing decision $t$, the active groups $\mathcal A_t\subseteq[M]$ expose
\begin{equation}
 \mathcal F_t=\bigcup_{j\in\mathcal A_t}\Pi(j),
 \qquad N_t^{\mathrm{active}}=M+|\mathcal F_t|.
 \label{eq:memory-interface}
\end{equation}
All Coarse entries remain active. Unselected Fine entries remain stored for later access but do not participate in attention.

\subsection{DART: Deformable Aggregation of Region-wise Tokens}
\label{sec:dart}
DART learns the \emph{capacity} (Fine-token count) and \emph{membership} of each of $M$ groups, which jointly define the units of compression and recall.

\begin{figure}[t]
  \centering
  \includegraphics[trim=9bp 4bp 13.5bp 7bp,clip,width=\linewidth]{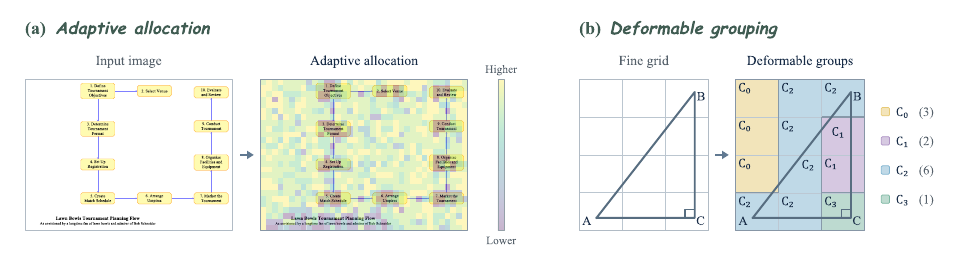}
  \figcaptionsep
  \caption{\textbf{DART adapts compression granularity and group shape to image content.} \textbf{(a)} Blank regions generally undergo stronger compression, while text-bearing regions retain finer granularity. Lower allocation-map values indicate fewer Fine tokens per Coarse token. \textbf{(b)} Deformable grouping follows the diagram's geometric structure rather than fixed grid boundaries: $C_0$ pools blank cells, $C_2$ covers much of the hypotenuse and part of the base, and $C_1$ captures the middle portion of the vertical leg. The cell containing the right-angle marker forms a separate group, $C_3$. Each group defines a Coarse token and its C2F entry; parentheses indicate the number of Fine members.}
  \label{fig:dart}
\end{figure}

\noindent\textbf{Allocate group sizes under a fixed total.}
Smaller groups provide finer representations and cheaper recall units (Fig.~\ref{fig:dart}(a)). An initial grid partitions $[N]$ into $M$ nonempty groups $\mathcal G_j^{(0)}$ with sizes $c_j^{(0)}$ and mean descriptors $\bar{\mathbf x}_j$. A learned score $a_j=\mathbf w_c^\top\bar{\mathbf x}_j$ adjusts their capacities:
\begin{equation}
 \widetilde c_j=1+(U-1)\operatorname{sigmoid}(b_j+a_j+\mu),
 \qquad \sum_{j=1}^{M}\widetilde c_j=N.
 \label{eq:capacity}
\end{equation}
Here $U>1$ bounds group size, and $b_j$ encodes the initial capacity. The shared offset $\mu$ enforces the fixed total; sum-preserving rounding yields integer capacities $c_j$ (details in Appendix~\ref{app:capacity-assignment}).

\noindent\textbf{Adapt membership to image content.}
DART scores membership by combining the grid prior $H_{ij}^{(0)}=\mathbf1_{\{i\in\mathcal G_j^{(0)}\}}$ with learned content affinity:
\begin{equation}
 S_{ij}=\gamma H_{ij}^{(0)}+
 \frac{(\mathbf W_{\mathrm F}\mathbf x_i^{\mathrm F})^\top
       (\mathbf W_{\mathrm C}\bar{\mathbf x}_j)}{\sqrt{d_a}}.
 \label{eq:affinity}
\end{equation}
The projections map tokens and regions to dimension $d_a$; $\gamma$ weights the prior. In confidence order, each Fine token enters its highest-scoring group with remaining capacity. The resulting groups partition $[N]$, satisfy $|\mathcal G_j|=c_j$, and can cross initial grid boundaries (Fig.~\ref{fig:dart}(b)).

\noindent\textbf{Share the compression and recall support.}
Within each group, normalized affinities form the pooling matrix $\mathbf W\in\mathbb R^{N\times M}$ and its C2F map:
\begin{equation}
 w_{ij}=\frac{\mathbf1_{\{i\in\mathcal G_j\}}\exp S_{ij}}
                  {\sum_{k\in\mathcal G_j}\exp S_{kj}},
 \qquad \mathbf X^{\mathrm C}=\mathbf W^\top\mathbf X^{\mathrm F},
 \qquad \Pi(j)=\mathcal G_j.
\label{eq:c2f}
\end{equation}
Thus compression and recall share support $\Pi(j)$, with Fine access count $|\mathcal F_t|=\sum_{j\in\mathcal A_t}c_j$.

The same weights construct the Coarse cache at decoder layer $\ell$:
\begin{equation}
 \mathbf K^{\mathrm C,\ell}=\mathbf W^\top\mathbf K^{\mathrm F,\ell},
 \qquad
 \mathbf V^{\mathrm C,\ell}=\mathbf W^\top E_\phi(\mathbf V^{\mathrm F,\ell}).
 \label{eq:coarse-kv}
\end{equation}
Keys are pooled directly and Values pass through the residual encoder $E_\phi$ before pooling. Recall uses the original Fine KV.

\subsection{TRACE: Temporal Routing for Adaptive Contextual Evidence}
\label{sec:trace}
At fixed token intervals, TRACE uses causally available backbone states and the previous active set to propose upcoming evidence and decide whether to retain or replace the current set.

\noindent\textbf{Represent decoding history.}
A learned combination and projection of hidden states across backbone layers forms $\mathbf u_t\in\mathbb R^{d_s}$. Inspired by selective state-space modeling~\citep{gu2023mamba}, $B$ recurrent branches retain decoding history at input-dependent timescales, starting from $\mathbf s_0^{(m)}=\mathbf0$:
\begin{equation}
 \mathbf s_t^{(m)}
 =\boldsymbol\rho_t^{(m)}\odot\mathbf s_{t-1}^{(m)}
   +(\mathbf1-\boldsymbol\rho_t^{(m)})\odot f(\mathbf u_t).
\label{eq:ssm}
\end{equation}
Here $\odot$ denotes elementwise multiplication, $\boldsymbol\rho_t^{(m)}\in(0,1)^{d_s}$ controls retention, and $f$ supplies new state content. A softmax mixture combines the branches as $\mathbf z_t=\sum_{m=1}^{B}\beta_t^{(m)}\mathbf s_t^{(m)}$. Appendix~\ref{app:history-analysis} gives the parameterization and history analysis.

\noindent\textbf{Propose evidence and decide when to replace it.}
Scaled dot-product scores $r_{t,j}$ match $\mathbf z_t$ to Coarse descriptors. A learned boundary $e(\mathbf z_t)$ yields membership probabilities $p_{t,j}=\operatorname{sigmoid}(r_{t,j}-e(\mathbf z_t))$ and the proposal $\widehat{\mathcal A}_t=\{j\in[M]:p_{t,j}>1/2\}$ (see Appendix~\ref{app:cache-interface}). Conditioned on the temporal state, the gate compares retained and proposed evidence and selects a HOLD or UPDATE action $g_t$:
\begin{equation}
 \mathcal A_t=\begin{cases}
 \mathcal A_{t-1},&g_t=\textsc{Hold},\\
 \widehat{\mathcal A}_t,&g_t=\textsc{Update}.
 \end{cases}
\label{eq:commit}
\end{equation}
HOLD preserves evidence; UPDATE can add, remove, or clear groups, allowing Coarse-only reasoning between Fine accesses. The recurrent state evolves under both actions. Appendix~\ref{app:cache-interface} defines the gate's evidence summaries.

\noindent\textbf{Attend to selected evidence with Joint-KV.}
For generation forward pass $n$, let $t(n)$ be the latest routing decision before that pass. Writing $\mathrm{KV}=(\mathbf K,\mathbf V)$, Joint-KV concatenates Coarse context, active Fine evidence, and available text history at each decoder layer:
\begin{equation}
 \mathrm{KV}_n^{\mathrm J}
 =[\mathrm{KV}^{\mathrm C};\,\mathrm{KV}^{\mathrm F}[\mathcal F_{t(n)}];\,\mathrm{KV}_n^{\mathrm T}].
\label{eq:joint-kv}
\end{equation}
During decoding after prefill, the frozen backbone attends jointly to these entries to compute the next-token distribution $p_{\theta_{\mathrm{Frozen}},\phi,\psi}(y_k\mid I,x,y_{<k})$. Algorithm~\ref{alg:fomer} summarizes the memory--access cycle.

\Needspace{4\baselineskip}
\subsection{Learning Compression and Evidence Access}
\label{sec:training}
With the backbone frozen, we first train DART parameters $\phi$, then fix DART and train TRACE parameters $\psi$ through supervised initialization and policy optimization.

\noindent\textbf{Distill compact visual memory.}
DART learns capacities, affinities, and Value aggregation from task-verified responses using token cross-entropy and KL distillation~\citep{hinton2015distilling} from full-visual teacher predictions under identical teacher-forced prefixes. Forward computation uses discrete groups; a capacity-constrained Sinkhorn surrogate supplies cross-group gradients~\citep{cuturi2013sinkhorn}. Appendices~\ref{app:capacity-assignment}--\ref{app:capacity-analysis} give the gradient construction and allocation analysis; Appendix~\ref{app:training-objectives} specifies the loss.

\noindent\textbf{Supervise upcoming evidence.}
TRACE observes only available hidden states and active-set history; future annotations supply supervision. Let $\mathcal E_s$ denote segment $s$'s teacher-derived Fine-token set. For the segment $s(t)$ aligned with the next token to be processed at routing time $t$, the target is
\begin{equation}
 \mathcal A_t^\star=\{j\in[M]:\Pi(j)\cap\mathcal E_{s(t)}\ne\varnothing\}.
 \label{eq:upcoming-target}
\end{equation}
Region BCE initializes set prediction; joint training adds gate cross-entropy, targeting UPDATE when the annotated set changes and HOLD otherwise.

\noindent\textbf{Optimize evidence access.}
TRACE is then optimized on student-generated trajectories using RLOO~\citep{ahmadian2024back} correctness feedback and success-conditioned feedback on mean Fine occupancy. Online teacher supervision guides upcoming region selections. Appendix~\ref{app:training-objectives} gives the complete objectives, occupancy definition, and action sampling.

%% file: sections/experiments.tex
\section{Experiments}
\label{sec:experiments}

\subsection{Experimental Setup}
\noindent\textbf{Benchmarks and baselines.}
We evaluate reasoning (Tab.~\ref{tab:main}) and general visual understanding (Tab.~\ref{tab:short-answer}); Appendix~\ref{app:benchmark-provenance} lists the benchmarks and prompting conventions. Full denotes the uncompressed backbone. The one-shot baselines are FastV~\citep{chen2024image}, VisionZip~\citep{yang2025visionzip}, and DivPrune~\citep{alvar2025divprune}. We also evaluate DSTP~\citep{kim2026and}, adapting its attention-change access rule to ViMoD's DART memory and frozen backbone.

\noindent\textbf{Model and configurations.}
We use a frozen \modelname{Qwen3-VL-4B-Instruct}~\citep{bai2025qwen3} backbone. ViMoD adds trainable modules containing only 0.0546\% of the backbone parameters. The 9:1 and 4:1 configurations retain nominal Coarse fractions of 11.1\% and 25\%, respectively. TRACE is trained only with the 9:1 compressor and reused directly with the separately trained 4:1 compressor.

\noindent\textbf{Training.}
Following Sec.~\ref{sec:training}, we train separate 4:1 and 9:1 DART compressors on the same verified response trajectories. \modelname{Qwen3.8-27B}~\citep{qwen38} provides TRACE's supervised annotations and online RL evidence; RL questions come from the supervised set. Training and evaluation sets are disjoint. Appendices~\ref{app:training-details} and~\ref{app:teacher-prompts} give the training recipe and annotation prompts, respectively.

\noindent\textbf{Visual budgets.}
The target budgets are 20\%, 30\%, and 40\% of the original visual-token count. Static methods retain the specified fraction. ViMoD counts persistent Coarse and active Fine tokens at each generation forward pass, averages within each example (Appendix~\ref{app:cost-analysis}, Eq.~(\ref{eq:visual-budget})), then averages examples equally. Fine tokens count throughout HOLD intervals. Calibration matches target average occupancy (see Appendix~\ref{app:budget-calibration} for calibration details); realized occupancy varies across examples and benchmarks. Measured occupancy and full inference costs are reported separately; Appendix~\ref{app:attention-accounting} defines the accounting.

\noindent\textbf{Evaluation.}
All methods use greedy decoding with zero presence penalty, with output limits matched across methods within each comparison. We use VLMEvalKit scoring~\citep{duan2024vlmevalkit}; Tabs.~\ref{tab:main} and~\ref{tab:short-answer} define normalization and averaging. Our decoding settings differ from the backbone's recommended configuration for visual tasks (see Appendix~\ref{app:generation-protocol} for the full protocol and discussion of comparability with reported backbone scores).

\subsection{Main Results}
\label{sec:accuracy}
\input{tables/reasoning_benchmarks}

\noindent\textbf{Reasoning under limited visual budgets.}
ViMoD leads the normalized reasoning average at every target budget and all eight benchmarks at 20\% budget (Tab.~\ref{tab:main}). Its average at 20\% is 39.0\% higher than the strongest one-shot baseline, DivPrune. Among the tested budgets, the best one-shot method \emph{requires 40\% to reach ViMoD's performance at 20\%}. The diminishing margin at larger budgets suggests that dynamic access is most valuable when a fixed context cannot cover shifting evidence needs (Figs.~\ref{fig:dynamic-demand}--\ref{fig:evidence-access}); a larger one-shot selection is more likely to include evidence needed later in the trajectory.

With the same 9:1 DART memory and frozen backbone, TRACE improves the normalized reasoning average over DSTP$^\dagger$ by 7.37--9.73 percentage points, supporting an access policy learned from decoding history and upcoming-evidence supervision.

\input{tables/general_benchmarks}

\noindent\textbf{General visual understanding and document--chart QA.}
ViMoD also leads the general-suite average at every budget, with its 20\% performance exceeding all one-shot results at 40\% (Tab.~\ref{tab:short-answer}). The smaller gains over one-shot methods than on reasoning tasks under tight budgets may reflect more stable visual requirements in brief responses, where a well-preserved initial representation can suffice; extended reasoning offers more scope for adaptive evidence access. Within this suite, gains are particularly pronounced on documents and charts. At 20\%, DocVQA and ChartQA improve by 19.59 and 28.16 percentage points over their respective strongest one-shot baselines. These gains may reflect DART's adaptive granularity: finer groups in text-bearing regions and stronger aggregation of blank regions help preserve compression-sensitive detail within a fixed Coarse-token budget (Fig.~\ref{fig:dart}(a)).

\noindent\textbf{Persistent Coarse context and Fine access.}
TRACE transfers directly from 9:1 to separately trained 4:1 memory without retraining, improving both suite averages at 30\% and 40\%. At both budgets, DocVQA favors 4:1, whereas MMMU-Pro favors 9:1. These preferences suggest a task-dependent balance between richer persistent Coarse context and the remaining Fine-access budget, which ViMoD accommodates with a shared policy across memory configurations.

\subsection{Ablation Studies}
\label{sec:ablations}
We ablate the 9:1 configuration on DynaMath and MMMU-Pro. All scores are accuracies (\%). The recall ablation uses the 30\% reference setting; realized occupancy may differ across variants.

\begin{table}[htbp]
\centering
\begin{minipage}[t]{0.475\linewidth}
\vspace{0pt}
\caption{Fine-recall accuracy (\%) with 9:1 compression at the 30\% reference budget.}\label{tab:recall-ablation}
\vspace{1pt}
\input{tables/recall_ablation}
\end{minipage}\hfill
\begin{minipage}[t]{0.475\linewidth}
\vspace{0pt}
\caption{Memory construction and RL ablations: accuracy (\%) at 9:1.}\label{tab:ablations}
\vspace{1pt}
\input{tables/component_ablation}
\end{minipage}
\end{table}
\noindent\textbf{Repeated evidence access.}
Repeated Fine recall outperforms both no recall and single recall (Tab.~\ref{tab:recall-ablation}). Relative to single recall, it improves DynaMath and MMMU-Pro by 3.35 and 8.12 percentage points, respectively. This pattern is consistent with evidence selected at one stage becoming insufficient as reasoning progresses (Fig.~\ref{fig:dynamic-demand}), supporting continued adaptation of the active evidence set.

\noindent\textbf{Learned visual memory.}
DART improves over mean pooling on regular $3\times3$ groups on both benchmarks at every budget (Tab.~\ref{tab:ablations}), with gains of 11.27 and 10.35 percentage points at 30\%. Fine recall remains enabled in both variants. The improvement therefore supports the value of learning how evidence is compressed and organized, even when the original Fine tokens remain recoverable.

\noindent\textbf{Policy optimization.}
Optimizing TRACE beyond supervised initialization improves accuracy at every budget, with gains of 5.99 and 4.22 percentage points on DynaMath and MMMU-Pro at 30\% (Tab.~\ref{tab:ablations}). These gains support complementing teacher-annotated access decisions with task feedback on the student's own reasoning trajectories.

\subsection{Efficiency Analysis}
\label{sec:efficiency}
\begin{figure}[htbp]
  \centering
  \includegraphics[trim=10bp 5.5bp 2.5bp 3bp,clip,width=0.80\linewidth]{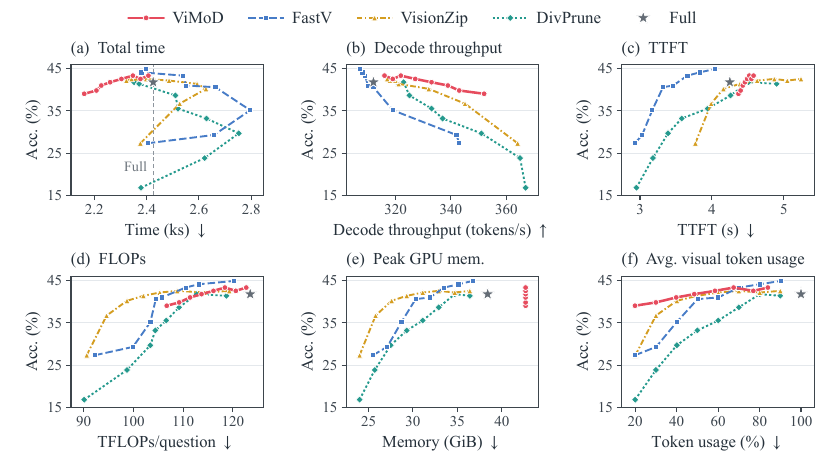}
  \figcaptionsep
  \caption{\textbf{Accuracy--efficiency trade-offs on MMMU-Pro.} Higher decoding throughput need not reduce total time. Lines connect eight target budgets (20--90\%) in increasing order; stars denote uncompressed Full, with its total time marked by the vertical guide in (a). Total time covers the complete run; decode throughput is generated tokens divided by decoding time; TTFT is time to first token. Visual occupancy averages visual-token usage over decoding forward passes relative to Full, counting Coarse and active Fine tokens for ViMoD.}
  \label{fig:efficiency-tradeoff}
\end{figure}

ViMoD reduces total inference time relative to Full across the evaluated 20--90\% budgets on MMMU-Pro (Fig.~\ref{fig:efficiency-tradeoff}).

\Needspace{5\baselineskip}
\noindent\textbf{Evaluation protocol.}
We compare ViMoD (9:1) with FastV~\citep{chen2024image}, VisionZip~\citep{yang2025visionzip}, and DivPrune~\citep{alvar2025divprune} on MMMU-Pro~\citep{yue2025mmmu} examples at target visual-token budgets of 20--90\% in increments of 10 percentage points. All methods use a single NVIDIA H200 GPU, \codeparam{batch\_size=8}, \codeparam{max\_new\_tokens=8192}, and the same inference backend. Fig.~\ref{fig:efficiency-tradeoff} reports accuracy--efficiency trade-offs; Appendix~\ref{app:efficiency-measurement} details measurement and FLOP accounting.

\noindent\textbf{End-to-end efficiency.}
At 40\%, ViMoD achieves higher accuracy than all three one-shot baselines with \emph{15.2--20.4\% less total time}. Although these baselines improve decoding throughput, they generate 15.9--29.9\% more tokens than Full, offsetting their per-token savings and increasing total time relative to Full, consistent with the response-level cost analysis in Fig.~\ref{fig:reasoning-cost}. ViMoD keeps output length close to Full, allowing its decoding gains to translate into lower total time.

Long output trajectories make autoregressive decoding dominant (94.4\% of Full's runtime in this evaluation). ViMoD retains full prefill and constructs Coarse memory, raising TTFT versus one-shot baselines at 20--50\%; these startup costs are amortized over subsequent decoding.

\input{tables/network_overhead}
ViMoD matches Full's accuracy at a 50\% target budget versus DivPrune's 80\%, using 39.5\% less measured visual occupancy and 3.9\% less total time.

\noindent
\textbf{Resource trade-offs.}\enspace
Retaining Fine and Coarse KV banks keeps peak memory fixed at 42.67\,GiB (Full: 38.40\,GiB): budgets regulate active access, while all evidence remains available for recall. DART and TRACE network computation takes 0.694\% of total time (Tab.~\ref{tab:efficiency-network-overhead}). Including initialization, memory processing, and dynamic-access overhead, selective Fine access still reduces total time versus Full across evaluated budgets.
\par

%% file: tables/reasoning_benchmarks.tex
\begin{table}[t]
\centering
\caption{\small{Mathematical, logical, and multidisciplinary reasoning. Parenthesized percentages are relative to Full; Avg. equally weights eight normalized scores. Bold and underlining mark the best and second-best scores within each target budget; $^\ddagger$ denotes explicit CoT prompting. ViMoD budgets specify average Coarse-plus-Fine occupancy. $^\dagger$ denotes our adaptation of DSTP's attention-change access rule to the same DART memory as ViMoD.}}
\label{tab:main}
\begin{mainresultstable}
\centering
\definecolor{resulthead}{gray}{0.95}
\definecolor{resultgroup}{gray}{0.30}
\setlength{\tabcolsep}{1.5pt}
\scalebox{0.6236}{%
\begin{tabularx}{632.4077pt}{l|*{7}{>{\hsize=.9742857143\hsize\linewidth=\hsize}C}>{\hsize=1.18\hsize\linewidth=\hsize}C|>{\kern2pt}r<{\kern2pt}}
\toprule\toprule
\rowcolor{resulthead} & \multicolumn{5}{c}{{\fontsize{9.5}{10.45}\selectfont\bfseries\color{black}Mathematics}} & \multicolumn{2}{c}{{\fontsize{9.5}{10.45}\selectfont\bfseries\color{black}Logic}} & \multicolumn{1}{c|}{{\fontsize{9.5}{10.45}\selectfont\bfseries\color{black}Multidisciplinary}} & \\
\noalign{\begingroup\color{resulthead}\hrule height\dimexpr\aboverulesep+\cmidrulewidth+\belowrulesep\relax\endgroup\kern-\dimexpr\aboverulesep+\cmidrulewidth+\belowrulesep\relax}
\cmidrule(lr){2-6}\cmidrule(lr){7-8}\cmidrule(lr){9-9}
\rowcolor{resulthead}\multirow{-2}{*}[2pt]{{\fontsize{9.5}{10.45}\selectfont\textbf{Method}}} & {\fontsize{9.3}{10.23}\selectfont \textbf{We-Math}$^\ddagger$} & {\fontsize{9.3}{10.23}\selectfont \textbf{DynaMath}$^\ddagger$} & {\fontsize{9.3}{10.23}\selectfont \textbf{MathVerse}$^\ddagger$} & {\fontsize{9.3}{10.23}\selectfont \textbf{MathVista}} & {\fontsize{9.3}{10.23}\selectfont \textbf{MathVision}} & {\fontsize{9.3}{10.23}\selectfont \textbf{LogicVista}} & {\fontsize{9.3}{10.23}\selectfont \textbf{VisualPuzzles}$^\ddagger$} & {\fontsize{9.3}{10.23}\selectfont \textbf{MMMU-Pro}$^\ddagger$} & \multirow{-2}{*}[2pt]{{\fontsize{9.5}{10.45}\selectfont\textbf{Avg. (\%)}}}\\
\midrule
Full (100\%) & 45.62{\fontsize{8}{8.8}\selectfont\color{ratio}(100.0\%)} & 61.88{\fontsize{8}{8.8}\selectfont\color{ratio}(100.0\%)} & 58.60{\fontsize{8}{8.8}\selectfont\color{ratio}(100.0\%)} & 72.80{\fontsize{8}{8.8}\selectfont\color{ratio}(100.0\%)} & 47.40{\fontsize{8}{8.8}\selectfont\color{ratio}(100.0\%)} & 37.36{\fontsize{8}{8.8}\selectfont\color{ratio}(100.0\%)} & 19.18{\fontsize{8}{8.8}\selectfont\color{ratio}(100.0\%)} & 30.92{\fontsize{8}{8.8}\selectfont\color{ratio}(100.0\%)} & 100.00\\
\midrule
\rowcolor{budgetband}\multicolumn{10}{c}{\itshape Retain {\color{budgetaccent}20\%} Visual Tokens}\\
FastV\, {\fontsize{7.5}{8.25}\selectfont (ECCV24)} & 16.95{\fontsize{8}{8.8}\selectfont\color{ratio}(37.2\%)} & 32.08{\fontsize{8}{8.8}\selectfont\color{ratio}(51.8\%)} & 37.31{\fontsize{8}{8.8}\selectfont\color{ratio}(63.7\%)} & 43.40{\fontsize{8}{8.8}\selectfont\color{ratio}(59.6\%)} & 35.56{\fontsize{8}{8.8}\selectfont\color{ratio}(75.0\%)} & 22.82{\fontsize{8}{8.8}\selectfont\color{ratio}(61.1\%)} & 11.47{\fontsize{8}{8.8}\selectfont\color{ratio}(59.8\%)} & 10.46{\fontsize{8}{8.8}\selectfont\color{ratio}(33.8\%)} & 55.25\\
VisionZip\, {\fontsize{7.5}{8.25}\selectfont (CVPR25)} & 17.43{\fontsize{8}{8.8}\selectfont\color{ratio}(38.2\%)} & 35.33{\fontsize{8}{8.8}\selectfont\color{ratio}(57.1\%)} & 37.79{\fontsize{8}{8.8}\selectfont\color{ratio}(64.5\%)} & 46.30{\fontsize{8}{8.8}\selectfont\color{ratio}(63.6\%)} & 33.91{\fontsize{8}{8.8}\selectfont\color{ratio}(71.5\%)} & 19.69{\fontsize{8}{8.8}\selectfont\color{ratio}(52.7\%)} & 8.99{\fontsize{8}{8.8}\selectfont\color{ratio}(46.9\%)} & 9.88{\fontsize{8}{8.8}\selectfont\color{ratio}(32.0\%)} & 53.31\\
DivPrune\, {\fontsize{7.5}{8.25}\selectfont (CVPR25)} & 23.52{\fontsize{8}{8.8}\selectfont\color{ratio}(51.6\%)} & 39.06{\fontsize{8}{8.8}\selectfont\color{ratio}(63.1\%)} & 41.75{\fontsize{8}{8.8}\selectfont\color{ratio}(71.2\%)} & 50.20{\fontsize{8}{8.8}\selectfont\color{ratio}(69.0\%)} & 35.36{\fontsize{8}{8.8}\selectfont\color{ratio}(74.6\%)} & 22.15{\fontsize{8}{8.8}\selectfont\color{ratio}(59.3\%)} & 8.90{\fontsize{8}{8.8}\selectfont\color{ratio}(46.4\%)} & 8.38{\fontsize{8}{8.8}\selectfont\color{ratio}(27.1\%)} & 57.78\\
DSTP$^\dagger$ (9:1)\, {\fontsize{7.5}{8.25}\selectfont (ECCV26)} & \cellcolor{second}\underline{34.89}{\fontsize{8}{8.8}\selectfont\color{ratio}(76.5\%)} & \cellcolor{second}\underline{48.20}{\fontsize{8}{8.8}\selectfont\color{ratio}(77.9\%)} & \cellcolor{second}\underline{49.21}{\fontsize{8}{8.8}\selectfont\color{ratio}(84.0\%)} & \cellcolor{second}\underline{61.00}{\fontsize{8}{8.8}\selectfont\color{ratio}(83.8\%)} & \cellcolor{second}\underline{38.82}{\fontsize{8}{8.8}\selectfont\color{ratio}(81.9\%)} & \cellcolor{second}\underline{26.85}{\fontsize{8}{8.8}\selectfont\color{ratio}(71.9\%)} & \cellcolor{second}\underline{12.93}{\fontsize{8}{8.8}\selectfont\color{ratio}(67.4\%)} & \cellcolor{second}\underline{12.49}{\fontsize{8}{8.8}\selectfont\color{ratio}(40.4\%)} & \cellcolor{second}\underline{72.96}\\
\textbf{ViMoD (9:1)} & \cellcolor{best}\textbf{37.24}{\fontsize{8}{8.8}\selectfont\color{ratio}(81.6\%)} & \cellcolor{best}\textbf{51.14}{\fontsize{8}{8.8}\selectfont\color{ratio}(82.6\%)} & \cellcolor{best}\textbf{50.25}{\fontsize{8}{8.8}\selectfont\color{ratio}(85.8\%)} & \cellcolor{best}\textbf{65.30}{\fontsize{8}{8.8}\selectfont\color{ratio}(89.7\%)} & \cellcolor{best}\textbf{40.43}{\fontsize{8}{8.8}\selectfont\color{ratio}(85.3\%)} & \cellcolor{best}\textbf{28.19}{\fontsize{8}{8.8}\selectfont\color{ratio}(75.5\%)} & \cellcolor{best}\textbf{15.15}{\fontsize{8}{8.8}\selectfont\color{ratio}(79.0\%)} & \cellcolor{best}\textbf{19.54}{\fontsize{8}{8.8}\selectfont\color{ratio}(63.2\%)} & \cellcolor{best}\textbf{80.33}\\
\midrule
\rowcolor{budgetband}\multicolumn{10}{c}{\itshape Retain {\color{budgetaccent}30\%} Visual Tokens}\\
FastV\, {\fontsize{7.5}{8.25}\selectfont (ECCV24)} & 22.76{\fontsize{8}{8.8}\selectfont\color{ratio}(49.9\%)} & 36.83{\fontsize{8}{8.8}\selectfont\color{ratio}(59.5\%)} & 41.95{\fontsize{8}{8.8}\selectfont\color{ratio}(71.6\%)} & 49.40{\fontsize{8}{8.8}\selectfont\color{ratio}(67.9\%)} & 36.18{\fontsize{8}{8.8}\selectfont\color{ratio}(76.3\%)} & 21.48{\fontsize{8}{8.8}\selectfont\color{ratio}(57.5\%)} & 11.22{\fontsize{8}{8.8}\selectfont\color{ratio}(58.5\%)} & 13.99{\fontsize{8}{8.8}\selectfont\color{ratio}(45.2\%)} & 60.80\\
VisionZip\, {\fontsize{7.5}{8.25}\selectfont (CVPR25)} & 28.10{\fontsize{8}{8.8}\selectfont\color{ratio}(61.6\%)} & 45.43{\fontsize{8}{8.8}\selectfont\color{ratio}(73.4\%)} & 46.62{\fontsize{8}{8.8}\selectfont\color{ratio}(79.6\%)} & 50.60{\fontsize{8}{8.8}\selectfont\color{ratio}(69.5\%)} & 37.50{\fontsize{8}{8.8}\selectfont\color{ratio}(79.1\%)} & 24.83{\fontsize{8}{8.8}\selectfont\color{ratio}(66.5\%)} & 12.07{\fontsize{8}{8.8}\selectfont\color{ratio}(62.9\%)} & 18.73{\fontsize{8}{8.8}\selectfont\color{ratio}(60.6\%)} & 69.14\\
DivPrune\, {\fontsize{7.5}{8.25}\selectfont (CVPR25)} & 29.52{\fontsize{8}{8.8}\selectfont\color{ratio}(64.7\%)} & 45.71{\fontsize{8}{8.8}\selectfont\color{ratio}(73.9\%)} & 46.42{\fontsize{8}{8.8}\selectfont\color{ratio}(79.2\%)} & 54.70{\fontsize{8}{8.8}\selectfont\color{ratio}(75.1\%)} & 36.48{\fontsize{8}{8.8}\selectfont\color{ratio}(77.0\%)} & 23.04{\fontsize{8}{8.8}\selectfont\color{ratio}(61.7\%)} & 11.99{\fontsize{8}{8.8}\selectfont\color{ratio}(62.5\%)} & 11.79{\fontsize{8}{8.8}\selectfont\color{ratio}(38.1\%)} & 66.53\\
DSTP$^\dagger$ (9:1)\, {\fontsize{7.5}{8.25}\selectfont (ECCV26)} & 35.86{\fontsize{8}{8.8}\selectfont\color{ratio}(78.6\%)} & 50.12{\fontsize{8}{8.8}\selectfont\color{ratio}(81.0\%)} & 51.22{\fontsize{8}{8.8}\selectfont\color{ratio}(87.4\%)} & 62.00{\fontsize{8}{8.8}\selectfont\color{ratio}(85.2\%)} & 38.75{\fontsize{8}{8.8}\selectfont\color{ratio}(81.8\%)} & 29.31{\fontsize{8}{8.8}\selectfont\color{ratio}(78.5\%)} & 14.12{\fontsize{8}{8.8}\selectfont\color{ratio}(73.6\%)} & 13.99{\fontsize{8}{8.8}\selectfont\color{ratio}(45.2\%)} & 76.41\\
DSTP$^\dagger$ (4:1)\, {\fontsize{7.5}{8.25}\selectfont (ECCV26)} & 37.70{\fontsize{8}{8.8}\selectfont\color{ratio}(82.6\%)} & \cellcolor{second}\underline{55.45}{\fontsize{8}{8.8}\selectfont\color{ratio}(89.6\%)} & 52.94{\fontsize{8}{8.8}\selectfont\color{ratio}(90.3\%)} & 64.80{\fontsize{8}{8.8}\selectfont\color{ratio}(89.0\%)} & 40.72{\fontsize{8}{8.8}\selectfont\color{ratio}(85.9\%)} & \cellcolor{second}\underline{31.10}{\fontsize{8}{8.8}\selectfont\color{ratio}(83.2\%)} & \cellcolor{second}\underline{16.18}{\fontsize{8}{8.8}\selectfont\color{ratio}(84.4\%)} & 21.33{\fontsize{8}{8.8}\selectfont\color{ratio}(69.0\%)} & 84.26\\
\textbf{ViMoD (9:1)} & \cellcolor{second}\underline{37.81}{\fontsize{8}{8.8}\selectfont\color{ratio}(82.9\%)} & 53.65{\fontsize{8}{8.8}\selectfont\color{ratio}(86.7\%)} & \cellcolor{second}\underline{53.12}{\fontsize{8}{8.8}\selectfont\color{ratio}(90.6\%)} & \cellcolor{best}\textbf{67.50}{\fontsize{8}{8.8}\selectfont\color{ratio}(92.7\%)} & \cellcolor{best}\textbf{41.71}{\fontsize{8}{8.8}\selectfont\color{ratio}(88.0\%)} & \cellcolor{second}\underline{31.10}{\fontsize{8}{8.8}\selectfont\color{ratio}(83.2\%)} & 16.10{\fontsize{8}{8.8}\selectfont\color{ratio}(83.9\%)} & \cellcolor{best}\textbf{25.03}{\fontsize{8}{8.8}\selectfont\color{ratio}(81.0\%)} & \cellcolor{second}\underline{86.14}\\
\textbf{ViMoD (4:1)} & \cellcolor{best}\textbf{40.29}{\fontsize{8}{8.8}\selectfont\color{ratio}(88.3\%)} & \cellcolor{best}\textbf{57.03}{\fontsize{8}{8.8}\selectfont\color{ratio}(92.2\%)} & \cellcolor{best}\textbf{53.55}{\fontsize{8}{8.8}\selectfont\color{ratio}(91.4\%)} & \cellcolor{second}\underline{67.10}{\fontsize{8}{8.8}\selectfont\color{ratio}(92.2\%)} & \cellcolor{second}\underline{41.18}{\fontsize{8}{8.8}\selectfont\color{ratio}(86.9\%)} & \cellcolor{best}\textbf{31.77}{\fontsize{8}{8.8}\selectfont\color{ratio}(85.0\%)} & \cellcolor{best}\textbf{16.87}{\fontsize{8}{8.8}\selectfont\color{ratio}(88.0\%)} & \cellcolor{second}\underline{24.05}{\fontsize{8}{8.8}\selectfont\color{ratio}(77.8\%)} & \cellcolor{best}\textbf{87.71}\\
\midrule
\rowcolor{budgetband}\multicolumn{10}{c}{\itshape Retain {\color{budgetaccent}40\%} Visual Tokens}\\
FastV\, {\fontsize{7.5}{8.25}\selectfont (ECCV24)} & 30.76{\fontsize{8}{8.8}\selectfont\color{ratio}(67.4\%)} & 41.54{\fontsize{8}{8.8}\selectfont\color{ratio}(67.1\%)} & 46.27{\fontsize{8}{8.8}\selectfont\color{ratio}(79.0\%)} & 50.70{\fontsize{8}{8.8}\selectfont\color{ratio}(69.6\%)} & 40.49{\fontsize{8}{8.8}\selectfont\color{ratio}(85.4\%)} & 24.83{\fontsize{8}{8.8}\selectfont\color{ratio}(66.5\%)} & 11.56{\fontsize{8}{8.8}\selectfont\color{ratio}(60.3\%)} & 18.09{\fontsize{8}{8.8}\selectfont\color{ratio}(58.5\%)} & 69.23\\
VisionZip\, {\fontsize{7.5}{8.25}\selectfont (CVPR25)} & 37.43{\fontsize{8}{8.8}\selectfont\color{ratio}(82.0\%)} & 52.55{\fontsize{8}{8.8}\selectfont\color{ratio}(84.9\%)} & 51.09{\fontsize{8}{8.8}\selectfont\color{ratio}(87.2\%)} & 57.40{\fontsize{8}{8.8}\selectfont\color{ratio}(78.8\%)} & \cellcolor{second}\underline{42.50}{\fontsize{8}{8.8}\selectfont\color{ratio}(89.7\%)} & 33.11{\fontsize{8}{8.8}\selectfont\color{ratio}(88.6\%)} & 14.38{\fontsize{8}{8.8}\selectfont\color{ratio}(75.0\%)} & 25.84{\fontsize{8}{8.8}\selectfont\color{ratio}(83.6\%)} & 83.73\\
DivPrune\, {\fontsize{7.5}{8.25}\selectfont (CVPR25)} & 34.38{\fontsize{8}{8.8}\selectfont\color{ratio}(75.4\%)} & 50.70{\fontsize{8}{8.8}\selectfont\color{ratio}(81.9\%)} & 50.51{\fontsize{8}{8.8}\selectfont\color{ratio}(86.2\%)} & 57.70{\fontsize{8}{8.8}\selectfont\color{ratio}(79.3\%)} & \cellcolor{best}\textbf{43.16}{\fontsize{8}{8.8}\selectfont\color{ratio}(91.1\%)} & 28.41{\fontsize{8}{8.8}\selectfont\color{ratio}(76.0\%)} & 11.82{\fontsize{8}{8.8}\selectfont\color{ratio}(61.6\%)} & 15.32{\fontsize{8}{8.8}\selectfont\color{ratio}(49.5\%)} & 75.13\\
DSTP$^\dagger$ (9:1)\, {\fontsize{7.5}{8.25}\selectfont (ECCV26)} & 37.24{\fontsize{8}{8.8}\selectfont\color{ratio}(81.6\%)} & 51.20{\fontsize{8}{8.8}\selectfont\color{ratio}(82.7\%)} & 52.01{\fontsize{8}{8.8}\selectfont\color{ratio}(88.8\%)} & 64.10{\fontsize{8}{8.8}\selectfont\color{ratio}(88.0\%)} & 39.90{\fontsize{8}{8.8}\selectfont\color{ratio}(84.2\%)} & 29.75{\fontsize{8}{8.8}\selectfont\color{ratio}(79.6\%)} & 15.75{\fontsize{8}{8.8}\selectfont\color{ratio}(82.1\%)} & 17.63{\fontsize{8}{8.8}\selectfont\color{ratio}(57.0\%)} & 80.51\\
DSTP$^\dagger$ (4:1)\, {\fontsize{7.5}{8.25}\selectfont (ECCV26)} & 39.81{\fontsize{8}{8.8}\selectfont\color{ratio}(87.3\%)} & \cellcolor{second}\underline{55.97}{\fontsize{8}{8.8}\selectfont\color{ratio}(90.4\%)} & 53.71{\fontsize{8}{8.8}\selectfont\color{ratio}(91.7\%)} & 66.10{\fontsize{8}{8.8}\selectfont\color{ratio}(90.8\%)} & 41.02{\fontsize{8}{8.8}\selectfont\color{ratio}(86.5\%)} & 31.54{\fontsize{8}{8.8}\selectfont\color{ratio}(84.4\%)} & \cellcolor{best}\textbf{17.38}{\fontsize{8}{8.8}\selectfont\color{ratio}(90.6\%)} & 21.27{\fontsize{8}{8.8}\selectfont\color{ratio}(68.8\%)} & 86.32\\
\textbf{ViMoD (9:1)} & \cellcolor{second}\underline{39.90}{\fontsize{8}{8.8}\selectfont\color{ratio}(87.5\%)} & 55.85{\fontsize{8}{8.8}\selectfont\color{ratio}(90.3\%)} & \cellcolor{best}\textbf{54.01}{\fontsize{8}{8.8}\selectfont\color{ratio}(92.2\%)} & \cellcolor{second}\underline{66.80}{\fontsize{8}{8.8}\selectfont\color{ratio}(91.8\%)} & 41.35{\fontsize{8}{8.8}\selectfont\color{ratio}(87.2\%)} & \cellcolor{second}\underline{33.33}{\fontsize{8}{8.8}\selectfont\color{ratio}(89.2\%)} & 15.92{\fontsize{8}{8.8}\selectfont\color{ratio}(83.0\%)} & \cellcolor{best}\textbf{27.69}{\fontsize{8}{8.8}\selectfont\color{ratio}(89.6\%)} & \cellcolor{second}\underline{88.83}\\
\textbf{ViMoD (4:1)} & \cellcolor{best}\textbf{41.43}{\fontsize{8}{8.8}\selectfont\color{ratio}(90.8\%)} & \cellcolor{best}\textbf{58.76}{\fontsize{8}{8.8}\selectfont\color{ratio}(95.0\%)} & \cellcolor{second}\underline{53.86}{\fontsize{8}{8.8}\selectfont\color{ratio}(91.9\%)} & \cellcolor{best}\textbf{67.50}{\fontsize{8}{8.8}\selectfont\color{ratio}(92.7\%)} & 42.04{\fontsize{8}{8.8}\selectfont\color{ratio}(88.7\%)} & \cellcolor{best}\textbf{34.90}{\fontsize{8}{8.8}\selectfont\color{ratio}(93.4\%)} & \cellcolor{second}\underline{17.12}{\fontsize{8}{8.8}\selectfont\color{ratio}(89.3\%)} & \cellcolor{second}\underline{26.82}{\fontsize{8}{8.8}\selectfont\color{ratio}(86.7\%)} & \cellcolor{best}\textbf{91.06}\\
\bottomrule\bottomrule
\end{tabularx}}
\end{mainresultstable}
\end{table}

%% file: tables/general_benchmarks.tex
\begin{table}[t]
\centering
\caption{General visual understanding and document--chart QA. Avg. equally weights ten Full-normalized metric columns. OCR EN/CN denote OCRBench v2 English/Chinese; \mbox{MME-P/C} report official perception/cognition scores. The DSTP adaptation ($^\dagger$), parenthesized percentages, and highlighting follow Tab.~\ref{tab:main}.}
\label{tab:short-answer}
\begin{mainresultstable}
\centering
\definecolor{resulthead}{gray}{0.95}
\definecolor{resultgroup}{gray}{0.30}
\setlength{\tabcolsep}{1pt}
\scalebox{0.6236}{%
\begin{tabularx}{632.4077pt}{l|>{\hsize=1.075\hsize\linewidth=\hsize}C>{\hsize=1.075\hsize\linewidth=\hsize}C>{\hsize=1.075\hsize\linewidth=\hsize}C>{\hsize=0.7\hsize\linewidth=\hsize}C>{\hsize=0.7\hsize\linewidth=\hsize}C>{\hsize=1.075\hsize\linewidth=\hsize}C>{\hsize=1.075\hsize\linewidth=\hsize}C>{\hsize=1.075\hsize\linewidth=\hsize}C>{\hsize=1.075\hsize\linewidth=\hsize}C>{\hsize=1.075\hsize\linewidth=\hsize}C|>{\kern2pt}r<{\kern2pt}}
\toprule\toprule
\rowcolor{resulthead} & \multicolumn{3}{c}{{\fontsize{9.5}{10.45}\selectfont\bfseries\color{black}Scene Understanding}} & \multicolumn{3}{c}{{\fontsize{9.5}{10.45}\selectfont\bfseries\color{black}General}} & \multicolumn{4}{c|}{{\fontsize{9.5}{10.45}\selectfont\bfseries\color{black}Documents and Charts}} & \\
\noalign{\begingroup\color{resulthead}\hrule height\dimexpr\aboverulesep+\cmidrulewidth+\belowrulesep\relax\endgroup\kern-\dimexpr\aboverulesep+\cmidrulewidth+\belowrulesep\relax}
\cmidrule(lr){2-4}\cmidrule(lr){5-7}\cmidrule(lr){8-11}
\rowcolor{resulthead}\multirow{-2}{*}[2pt]{{\fontsize{9.5}{10.45}\selectfont\textbf{Method}}} & {\fontsize{9.3}{10.23}\selectfont \textbf{GQA}} & {\fontsize{9.3}{10.23}\selectfont \textbf{POPE}} & {\fontsize{9.3}{10.23}\selectfont \textbf{V$^*$}} & {\fontsize{9.3}{10.23}\selectfont \textbf{MME-P}} & {\fontsize{9.3}{10.23}\selectfont \textbf{MME-C}} & {\fontsize{9.3}{10.23}\selectfont \textbf{MMBench}} & {\fontsize{9.3}{10.23}\selectfont \textbf{DocVQA}} & {\fontsize{9.3}{10.23}\selectfont \textbf{OCR EN}} & {\fontsize{9.3}{10.23}\selectfont \textbf{OCR CN}} & {\fontsize{9.3}{10.23}\selectfont \textbf{ChartQA}} & \multirow{-2}{*}[2pt]{{\fontsize{9.5}{10.45}\selectfont\textbf{Avg. (\%)}}}\\
\midrule
Full (100\%) & 62.11{\fontsize{8}{8.8}\selectfont\color{ratio}(100.0\%)} & 89.89{\fontsize{8}{8.8}\selectfont\color{ratio}(100.0\%)} & 68.59{\fontsize{8}{8.8}\selectfont\color{ratio}(100.0\%)} & 1693.06 & 610.71 & 83.51{\fontsize{8}{8.8}\selectfont\color{ratio}(100.0\%)} & 91.68{\fontsize{8}{8.8}\selectfont\color{ratio}(100.0\%)} & 44.89{\fontsize{8}{8.8}\selectfont\color{ratio}(100.0\%)} & 41.40{\fontsize{8}{8.8}\selectfont\color{ratio}(100.0\%)} & 80.64{\fontsize{8}{8.8}\selectfont\color{ratio}(100.0\%)} & 100.00\\
\midrule
\rowcolor{budgetband}\multicolumn{12}{c}{\itshape Retain {\color{budgetaccent}20\%} Visual Tokens}\\
FastV\, {\fontsize{7.5}{8.25}\selectfont (ECCV24)} & 50.00{\fontsize{8}{8.8}\selectfont\color{ratio}(80.5\%)} & 77.24{\fontsize{8}{8.8}\selectfont\color{ratio}(85.9\%)} & 56.54{\fontsize{8}{8.8}\selectfont\color{ratio}(82.4\%)} & 1346.63 & 312.50 & 74.14{\fontsize{8}{8.8}\selectfont\color{ratio}(88.8\%)} & 36.14{\fontsize{8}{8.8}\selectfont\color{ratio}(39.4\%)} & 29.74{\fontsize{8}{8.8}\selectfont\color{ratio}(66.3\%)} & 19.87{\fontsize{8}{8.8}\selectfont\color{ratio}(48.0\%)} & 30.84{\fontsize{8}{8.8}\selectfont\color{ratio}(38.2\%)} & 66.03\\
VisionZip\, {\fontsize{7.5}{8.25}\selectfont (CVPR25)} & 56.41{\fontsize{8}{8.8}\selectfont\color{ratio}(90.8\%)} & 84.37{\fontsize{8}{8.8}\selectfont\color{ratio}(93.9\%)} & 59.16{\fontsize{8}{8.8}\selectfont\color{ratio}(86.3\%)} & 1423.19 & 373.57 & 74.91{\fontsize{8}{8.8}\selectfont\color{ratio}(89.7\%)} & 47.99{\fontsize{8}{8.8}\selectfont\color{ratio}(52.3\%)} & 28.12{\fontsize{8}{8.8}\selectfont\color{ratio}(62.6\%)} & 18.51{\fontsize{8}{8.8}\selectfont\color{ratio}(44.7\%)} & 48.44{\fontsize{8}{8.8}\selectfont\color{ratio}(60.1\%)} & 72.56\\
DivPrune\, {\fontsize{7.5}{8.25}\selectfont (CVPR25)} & 58.90{\fontsize{8}{8.8}\selectfont\color{ratio}(94.8\%)} & 88.10{\fontsize{8}{8.8}\selectfont\color{ratio}(98.0\%)} & 59.69{\fontsize{8}{8.8}\selectfont\color{ratio}(87.0\%)} & \cellcolor{second}\underline{1620.48} & \cellcolor{second}\underline{415.71} & 78.87{\fontsize{8}{8.8}\selectfont\color{ratio}(94.4\%)} & 56.13{\fontsize{8}{8.8}\selectfont\color{ratio}(61.2\%)} & 33.23{\fontsize{8}{8.8}\selectfont\color{ratio}(74.0\%)} & 24.06{\fontsize{8}{8.8}\selectfont\color{ratio}(58.1\%)} & 44.64{\fontsize{8}{8.8}\selectfont\color{ratio}(55.4\%)} & 78.68\\
DSTP$^\dagger$ (9:1)\, {\fontsize{7.5}{8.25}\selectfont (ECCV26)} & \cellcolor{second}\underline{61.46}{\fontsize{8}{8.8}\selectfont\color{ratio}(99.0\%)} & \cellcolor{second}\underline{89.82}{\fontsize{8}{8.8}\selectfont\color{ratio}(99.9\%)} & \cellcolor{second}\underline{64.92}{\fontsize{8}{8.8}\selectfont\color{ratio}(94.6\%)} & \cellcolor{best}\textbf{1671.11} & \cellcolor{best}\textbf{610.00} & \cellcolor{best}\textbf{84.28}{\fontsize{8}{8.8}\selectfont\color{ratio}(100.9\%)} & \cellcolor{second}\underline{71.62}{\fontsize{8}{8.8}\selectfont\color{ratio}(78.1\%)} & \cellcolor{second}\underline{40.66}{\fontsize{8}{8.8}\selectfont\color{ratio}(90.6\%)} & \cellcolor{second}\underline{26.40}{\fontsize{8}{8.8}\selectfont\color{ratio}(63.8\%)} & \cellcolor{second}\underline{74.92}{\fontsize{8}{8.8}\selectfont\color{ratio}(92.9\%)} & \cellcolor{second}\underline{91.84}\\
\textbf{ViMoD (9:1)} & \cellcolor{best}\textbf{61.65}{\fontsize{8}{8.8}\selectfont\color{ratio}(99.3\%)} & \cellcolor{best}\textbf{89.84}{\fontsize{8}{8.8}\selectfont\color{ratio}(99.9\%)} & \cellcolor{best}\textbf{65.45}{\fontsize{8}{8.8}\selectfont\color{ratio}(95.4\%)} & \cellcolor{best}\textbf{1671.11} & \cellcolor{best}\textbf{610.00} & \cellcolor{second}\underline{84.11}{\fontsize{8}{8.8}\selectfont\color{ratio}(100.7\%)} & \cellcolor{best}\textbf{75.72}{\fontsize{8}{8.8}\selectfont\color{ratio}(82.6\%)} & \cellcolor{best}\textbf{42.57}{\fontsize{8}{8.8}\selectfont\color{ratio}(94.8\%)} & \cellcolor{best}\textbf{29.56}{\fontsize{8}{8.8}\selectfont\color{ratio}(71.4\%)} & \cellcolor{best}\textbf{76.60}{\fontsize{8}{8.8}\selectfont\color{ratio}(95.0\%)} & \cellcolor{best}\textbf{93.77}\\
\midrule
\rowcolor{budgetband}\multicolumn{12}{c}{\itshape Retain {\color{budgetaccent}30\%} Visual Tokens}\\
FastV\, {\fontsize{7.5}{8.25}\selectfont (ECCV24)} & 54.09{\fontsize{8}{8.8}\selectfont\color{ratio}(87.1\%)} & 81.90{\fontsize{8}{8.8}\selectfont\color{ratio}(91.1\%)} & 58.64{\fontsize{8}{8.8}\selectfont\color{ratio}(85.5\%)} & 1460.27 & 369.29 & 78.09{\fontsize{8}{8.8}\selectfont\color{ratio}(93.5\%)} & 49.30{\fontsize{8}{8.8}\selectfont\color{ratio}(53.8\%)} & 36.00{\fontsize{8}{8.8}\selectfont\color{ratio}(80.2\%)} & 23.65{\fontsize{8}{8.8}\selectfont\color{ratio}(57.1\%)} & 45.04{\fontsize{8}{8.8}\selectfont\color{ratio}(55.9\%)} & 75.09\\
VisionZip\, {\fontsize{7.5}{8.25}\selectfont (CVPR25)} & 59.05{\fontsize{8}{8.8}\selectfont\color{ratio}(95.1\%)} & 88.21{\fontsize{8}{8.8}\selectfont\color{ratio}(98.1\%)} & 64.40{\fontsize{8}{8.8}\selectfont\color{ratio}(93.9\%)} & 1628.01 & 383.57 & 78.35{\fontsize{8}{8.8}\selectfont\color{ratio}(93.8\%)} & 68.84{\fontsize{8}{8.8}\selectfont\color{ratio}(75.1\%)} & 37.79{\fontsize{8}{8.8}\selectfont\color{ratio}(84.2\%)} & 26.59{\fontsize{8}{8.8}\selectfont\color{ratio}(64.2\%)} & 66.56{\fontsize{8}{8.8}\selectfont\color{ratio}(82.5\%)} & 84.59\\
DivPrune\, {\fontsize{7.5}{8.25}\selectfont (CVPR25)} & 60.40{\fontsize{8}{8.8}\selectfont\color{ratio}(97.2\%)} & 89.01{\fontsize{8}{8.8}\selectfont\color{ratio}(99.0\%)} & 63.35{\fontsize{8}{8.8}\selectfont\color{ratio}(92.4\%)} & 1670.39 & 527.14 & 80.24{\fontsize{8}{8.8}\selectfont\color{ratio}(96.1\%)} & 68.33{\fontsize{8}{8.8}\selectfont\color{ratio}(74.5\%)} & 38.64{\fontsize{8}{8.8}\selectfont\color{ratio}(86.1\%)} & 28.96{\fontsize{8}{8.8}\selectfont\color{ratio}(70.0\%)} & 56.72{\fontsize{8}{8.8}\selectfont\color{ratio}(70.3\%)} & 87.06\\
DSTP$^\dagger$ (9:1)\, {\fontsize{7.5}{8.25}\selectfont (ECCV26)} & 61.58{\fontsize{8}{8.8}\selectfont\color{ratio}(99.1\%)} & \cellcolor{best}\textbf{89.84}{\fontsize{8}{8.8}\selectfont\color{ratio}(99.9\%)} & \cellcolor{second}\underline{65.97}{\fontsize{8}{8.8}\selectfont\color{ratio}(96.2\%)} & \cellcolor{best}\textbf{1679.40} & 607.86 & \cellcolor{second}\underline{84.11}{\fontsize{8}{8.8}\selectfont\color{ratio}(100.7\%)} & 76.86{\fontsize{8}{8.8}\selectfont\color{ratio}(83.8\%)} & 45.19{\fontsize{8}{8.8}\selectfont\color{ratio}(100.7\%)} & 31.60{\fontsize{8}{8.8}\selectfont\color{ratio}(76.3\%)} & 76.64{\fontsize{8}{8.8}\selectfont\color{ratio}(95.0\%)} & 95.06\\
DSTP$^\dagger$ (4:1)\, {\fontsize{7.5}{8.25}\selectfont (ECCV26)} & 61.61{\fontsize{8}{8.8}\selectfont\color{ratio}(99.2\%)} & 89.79{\fontsize{8}{8.8}\selectfont\color{ratio}(99.9\%)} & \cellcolor{best}\textbf{66.49}{\fontsize{8}{8.8}\selectfont\color{ratio}(96.9\%)} & 1673.42 & \cellcolor{best}\textbf{612.14} & \cellcolor{second}\underline{84.11}{\fontsize{8}{8.8}\selectfont\color{ratio}(100.7\%)} & \cellcolor{second}\underline{84.89}{\fontsize{8}{8.8}\selectfont\color{ratio}(92.6\%)} & \cellcolor{second}\underline{46.86}{\fontsize{8}{8.8}\selectfont\color{ratio}(104.4\%)} & \cellcolor{second}\underline{33.54}{\fontsize{8}{8.8}\selectfont\color{ratio}(81.0\%)} & \cellcolor{best}\textbf{80.36}{\fontsize{8}{8.8}\selectfont\color{ratio}(99.7\%)} & \cellcolor{second}\underline{97.35}\\
\textbf{ViMoD (9:1)} & \cellcolor{second}\underline{61.64}{\fontsize{8}{8.8}\selectfont\color{ratio}(99.2\%)} & \cellcolor{second}\underline{89.82}{\fontsize{8}{8.8}\selectfont\color{ratio}(99.9\%)} & 65.45{\fontsize{8}{8.8}\selectfont\color{ratio}(95.4\%)} & \cellcolor{second}\underline{1675.63} & \cellcolor{second}\underline{610.00} & \cellcolor{best}\textbf{84.19}{\fontsize{8}{8.8}\selectfont\color{ratio}(100.8\%)} & 81.03{\fontsize{8}{8.8}\selectfont\color{ratio}(88.4\%)} & 45.72{\fontsize{8}{8.8}\selectfont\color{ratio}(101.8\%)} & 32.19{\fontsize{8}{8.8}\selectfont\color{ratio}(77.8\%)} & 78.60{\fontsize{8}{8.8}\selectfont\color{ratio}(97.5\%)} & 95.97\\
\textbf{ViMoD (4:1)} & \cellcolor{best}\textbf{61.65}{\fontsize{8}{8.8}\selectfont\color{ratio}(99.3\%)} & 89.73{\fontsize{8}{8.8}\selectfont\color{ratio}(99.8\%)} & 64.92{\fontsize{8}{8.8}\selectfont\color{ratio}(94.6\%)} & 1673.42 & \cellcolor{best}\textbf{612.14} & \cellcolor{second}\underline{84.11}{\fontsize{8}{8.8}\selectfont\color{ratio}(100.7\%)} & \cellcolor{best}\textbf{85.79}{\fontsize{8}{8.8}\selectfont\color{ratio}(93.6\%)} & \cellcolor{best}\textbf{47.12}{\fontsize{8}{8.8}\selectfont\color{ratio}(105.0\%)} & \cellcolor{best}\textbf{34.01}{\fontsize{8}{8.8}\selectfont\color{ratio}(82.1\%)} & \cellcolor{second}\underline{80.28}{\fontsize{8}{8.8}\selectfont\color{ratio}(99.6\%)} & \cellcolor{best}\textbf{97.38}\\
\midrule
\rowcolor{budgetband}\multicolumn{12}{c}{\itshape Retain {\color{budgetaccent}40\%} Visual Tokens}\\
FastV\, {\fontsize{7.5}{8.25}\selectfont (ECCV24)} & 56.63{\fontsize{8}{8.8}\selectfont\color{ratio}(91.2\%)} & 84.72{\fontsize{8}{8.8}\selectfont\color{ratio}(94.2\%)} & 62.30{\fontsize{8}{8.8}\selectfont\color{ratio}(90.8\%)} & 1578.74 & 371.43 & 80.07{\fontsize{8}{8.8}\selectfont\color{ratio}(95.9\%)} & 62.02{\fontsize{8}{8.8}\selectfont\color{ratio}(67.6\%)} & 41.30{\fontsize{8}{8.8}\selectfont\color{ratio}(92.0\%)} & 28.07{\fontsize{8}{8.8}\selectfont\color{ratio}(67.8\%)} & 58.48{\fontsize{8}{8.8}\selectfont\color{ratio}(72.5\%)} & 82.62\\
VisionZip\, {\fontsize{7.5}{8.25}\selectfont (CVPR25)} & 60.67{\fontsize{8}{8.8}\selectfont\color{ratio}(97.7\%)} & 89.54{\fontsize{8}{8.8}\selectfont\color{ratio}(99.6\%)} & 65.45{\fontsize{8}{8.8}\selectfont\color{ratio}(95.4\%)} & 1645.17 & 447.50 & 80.67{\fontsize{8}{8.8}\selectfont\color{ratio}(96.6\%)} & 79.97{\fontsize{8}{8.8}\selectfont\color{ratio}(87.2\%)} & 44.86{\fontsize{8}{8.8}\selectfont\color{ratio}(99.9\%)} & 32.70{\fontsize{8}{8.8}\selectfont\color{ratio}(79.0\%)} & 73.84{\fontsize{8}{8.8}\selectfont\color{ratio}(91.6\%)} & 91.75\\
DivPrune\, {\fontsize{7.5}{8.25}\selectfont (CVPR25)} & 60.80{\fontsize{8}{8.8}\selectfont\color{ratio}(97.9\%)} & 89.10{\fontsize{8}{8.8}\selectfont\color{ratio}(99.1\%)} & \cellcolor{second}\underline{65.97}{\fontsize{8}{8.8}\selectfont\color{ratio}(96.2\%)} & \cellcolor{best}\textbf{1715.35} & 541.43 & 81.36{\fontsize{8}{8.8}\selectfont\color{ratio}(97.4\%)} & 76.16{\fontsize{8}{8.8}\selectfont\color{ratio}(83.1\%)} & 43.00{\fontsize{8}{8.8}\selectfont\color{ratio}(95.8\%)} & 32.83{\fontsize{8}{8.8}\selectfont\color{ratio}(79.3\%)} & 65.60{\fontsize{8}{8.8}\selectfont\color{ratio}(81.3\%)} & 92.01\\
DSTP$^\dagger$ (9:1)\, {\fontsize{7.5}{8.25}\selectfont (ECCV26)} & \cellcolor{second}\underline{61.78}{\fontsize{8}{8.8}\selectfont\color{ratio}(99.5\%)} & 89.76{\fontsize{8}{8.8}\selectfont\color{ratio}(99.9\%)} & \cellcolor{second}\underline{65.97}{\fontsize{8}{8.8}\selectfont\color{ratio}(96.2\%)} & \cellcolor{second}\underline{1679.24} & 607.86 & \cellcolor{second}\underline{84.11}{\fontsize{8}{8.8}\selectfont\color{ratio}(100.7\%)} & 81.83{\fontsize{8}{8.8}\selectfont\color{ratio}(89.3\%)} & 46.33{\fontsize{8}{8.8}\selectfont\color{ratio}(103.2\%)} & 33.89{\fontsize{8}{8.8}\selectfont\color{ratio}(81.9\%)} & 78.44{\fontsize{8}{8.8}\selectfont\color{ratio}(97.3\%)} & 96.65\\
DSTP$^\dagger$ (4:1)\, {\fontsize{7.5}{8.25}\selectfont (ECCV26)} & 61.77{\fontsize{8}{8.8}\selectfont\color{ratio}(99.5\%)} & \cellcolor{best}\textbf{89.86}{\fontsize{8}{8.8}\selectfont\color{ratio}(100.0\%)} & \cellcolor{best}\textbf{66.49}{\fontsize{8}{8.8}\selectfont\color{ratio}(96.9\%)} & 1675.67 & \cellcolor{second}\underline{610.00} & \cellcolor{second}\underline{84.11}{\fontsize{8}{8.8}\selectfont\color{ratio}(100.7\%)} & \cellcolor{second}\underline{85.28}{\fontsize{8}{8.8}\selectfont\color{ratio}(93.0\%)} & \cellcolor{second}\underline{48.57}{\fontsize{8}{8.8}\selectfont\color{ratio}(108.2\%)} & \cellcolor{second}\underline{35.08}{\fontsize{8}{8.8}\selectfont\color{ratio}(84.7\%)} & \cellcolor{best}\textbf{81.28}{\fontsize{8}{8.8}\selectfont\color{ratio}(100.8\%)} & \cellcolor{second}\underline{98.27}\\
\textbf{ViMoD (9:1)} & 61.71{\fontsize{8}{8.8}\selectfont\color{ratio}(99.4\%)} & \cellcolor{second}\underline{89.83}{\fontsize{8}{8.8}\selectfont\color{ratio}(99.9\%)} & \cellcolor{second}\underline{65.97}{\fontsize{8}{8.8}\selectfont\color{ratio}(96.2\%)} & 1670.38 & \cellcolor{second}\underline{610.00} & 83.93{\fontsize{8}{8.8}\selectfont\color{ratio}(100.5\%)} & 84.69{\fontsize{8}{8.8}\selectfont\color{ratio}(92.4\%)} & 48.08{\fontsize{8}{8.8}\selectfont\color{ratio}(107.1\%)} & 34.44{\fontsize{8}{8.8}\selectfont\color{ratio}(83.2\%)} & 79.36{\fontsize{8}{8.8}\selectfont\color{ratio}(98.4\%)} & 97.56\\
\textbf{ViMoD (4:1)} & \cellcolor{best}\textbf{61.93}{\fontsize{8}{8.8}\selectfont\color{ratio}(99.7\%)} & 89.80{\fontsize{8}{8.8}\selectfont\color{ratio}(99.9\%)} & 65.45{\fontsize{8}{8.8}\selectfont\color{ratio}(95.4\%)} & 1674.17 & \cellcolor{best}\textbf{619.64} & \cellcolor{best}\textbf{84.19}{\fontsize{8}{8.8}\selectfont\color{ratio}(100.8\%)} & \cellcolor{best}\textbf{88.04}{\fontsize{8}{8.8}\selectfont\color{ratio}(96.0\%)} & \cellcolor{best}\textbf{48.79}{\fontsize{8}{8.8}\selectfont\color{ratio}(108.7\%)} & \cellcolor{best}\textbf{36.14}{\fontsize{8}{8.8}\selectfont\color{ratio}(87.3\%)} & \cellcolor{second}\underline{80.92}{\fontsize{8}{8.8}\selectfont\color{ratio}(100.3\%)} & \cellcolor{best}\textbf{98.86}\\
\bottomrule\bottomrule
\end{tabularx}}
\end{mainresultstable}
\end{table}

%% file: tables/recall_ablation.tex
\begin{ablationtable}
\fontsize{7.4}{8.1}\selectfont
\renewcommand{\arraystretch}{1.12}
\setlength{\aboverulesep}{0.8pt}
\setlength{\belowrulesep}{1.2pt}
\begin{tabular*}{\linewidth}{@{\extracolsep{\fill}}lcc}
\toprule\toprule
\textbf{Setting} & \textbf{DynaMath} & \textbf{MMMU-Pro}\\
\midrule
ViMoD & \textbf{53.65} & \textbf{25.03}\\
\midrule
w/o Fine recall & 47.72 & 12.20\\
w/ single recall & 50.30 & 16.91\\
\bottomrule\bottomrule
\end{tabular*}
\end{ablationtable}

%% file: tables/component_ablation.tex
\begin{ablationtable}[1.6pt]
\fontsize{7.4}{8.1}\selectfont
\renewcommand{\arraystretch}{0.94}
\setlength{\aboverulesep}{0.8pt}
\setlength{\belowrulesep}{1.2pt}
\begin{tabular*}{\linewidth}{@{\extracolsep{\fill}}l*{6}{c}}
\toprule\toprule
\multirow{2}{*}[-2pt]{\textbf{Setting}} & \multicolumn{3}{c}{\textbf{DynaMath}} & \multicolumn{3}{c}{\textbf{MMMU-Pro}}\\
\cmidrule(lr){2-4}\cmidrule(lr){5-7}
& \abudget{20} & \abudget{30} & \abudget{40} & \abudget{20} & \abudget{30} & \abudget{40}\\
\midrule
ViMoD & \textbf{51.14} & \textbf{53.65} & \textbf{55.85} & \textbf{19.54} & \textbf{25.03} & \textbf{27.69}\\
\midrule
w/ mean pooling & 39.14 & 42.38 & 45.95 & \phantom{0}8.32 & 14.68 & 18.50\\
w/o RL (SFT only) & 47.66 & 47.66 & 47.66 & 15.95 & 20.81 & 23.82\\
\bottomrule\bottomrule
\end{tabular*}
\end{ablationtable}

%% file: tables/network_overhead.tex
\begin{wraptable}{R}{0.35\textwidth}
  \vspace{-0pt}
  \vspace{-\abovecaptionskip}
  \centering
  \caption{Network runtime (\% of total), averaged over eight budgets.}
  \label{tab:efficiency-network-overhead}
  \vspace{1pt}
  \begin{ablationtable}[3pt]
  \begin{tabularx}{0.86\linewidth}{@{}*{2}{>{\centering\arraybackslash}X}@{}}
    \toprule\toprule
    \textbf{DART} & \textbf{TRACE} \\
    \midrule
    0.078\% & 0.616\% \\
    \bottomrule\bottomrule
  \end{tabularx}
  \end{ablationtable}
\end{wraptable}

%% file: sections/limitations.tex
\section{Limitations}
\label{sec:limitations}
Constructing Coarse memory adds initialization work, and retaining both Coarse and Fine KV increases storage requirements. Short responses or sustained access to broad visual evidence may limit the decoding savings available to offset initialization and dynamic access overhead. C2F granularity, fixed routing intervals, and teacher supervision constrain evidence access: recall can activate unnecessary group members, while missed groups or delayed updates can leave useful evidence inaccessible. Generalization across backbones and robustness to alternative decoding configurations remain unverified.

%% file: sections/conclusion.tex
\section{Conclusion}
\label{sec:conclusion}
ViMoD enables efficient multi-step visual reasoning through content-adaptive memory and selective evidence access. DART organizes recoverable visual detail, while TRACE learns to anticipate evidence needs from decoding history. This design improves accuracy under tight visual budgets and reduces total inference time in our evaluation, showing the value of coordinating memory organization with access throughout reasoning.

%% file: appendices/additional_method.tex
\clearpage
\section{Method Details and Derivations}
\label{app:method}
We derive the capacity-allocation and temporal-memory properties, specify the discrete grouping surrogate and learning objectives, and summarize the evidence-access algorithm.

\begin{center}
\small
\begin{tabular}{@{}ll@{}}
\toprule
Notation & Meaning \\
\midrule
$\mathbf X^{\mathrm F}\in\mathbb R^{N\times d},\ \mathbf X^{\mathrm C}\in\mathbb R^{M\times d}$ & Fine and Coarse feature matrices \\
$\mathbf W\in\mathbb R^{N\times M},\ \Pi(j)=\mathcal G_j$ & Pooling weights and group-to-Fine index map \\
$\widetilde c_j,\ c_j,\ U$ & Relaxed group size, integer group size, upper bound \\
$\mathbf K^{\mathrm F},\mathbf V^{\mathrm F},\mathbf K^{\mathrm C},\mathbf V^{\mathrm C}$ & Fine and Coarse caches \\
$\mathcal A_t,\ \mathcal F_t,\ g_t$ & Active groups, active Fine indices, gate action \\
$\mathbf s_t^{(m)},\ \mathbf z_t$ & Branch state and mixed temporal state \\
$n,\ t(n),\ T_i$ & Response-generation forward pass, latest routing step, count \\
$C_i$ & Mean Fine occupancy during response generation \\
$\theta_{\mathrm{Frozen}},\ \phi,\ \psi$ & Frozen backbone parameters, DART parameters, TRACE parameters \\
\bottomrule
\end{tabular}
\end{center}

\subsection{DART: Memory Construction}
\subsubsection{Capacity-Constrained Assignment and Pooling}
\label{app:capacity-assignment}
\paragraph{Feasible capacities and integer rounding.}
Let $c_j^{(0)}$ be the number of Fine tokens in initial grid region $j$. For an integer upper bound $U>1$, the capacity prior is
\begin{equation}
 \eta_j=\min\!\left\{1-\epsilon_b,\,
 \max\!\left\{\epsilon_b,\frac{c_j^{(0)}-1}{U-1}\right\}\right\},
 \qquad b_j=\log\frac{\eta_j}{1-\eta_j},
 \label{eq:capacity-prior}
\end{equation}
where $0<\epsilon_b<\tfrac12$ keeps the prior finite at either bound. The constraint in Equation~(\ref{eq:capacity}) is feasible for $M\leq N\leq MU$. Taking floors and assigning the remaining slots to the largest fractional remainders produces integer capacities satisfying $1\leq c_j\leq U$ and $\sum_jc_j=N$.

\paragraph{Discrete assignment.}
Fine tokens are processed in descending assignment confidence, defined by the gap between their largest and second-largest affinity scores. Each token enters its highest-scoring group with remaining capacity. Total capacity equals $N$, so this procedure assigns every token and fills every group.

\paragraph{Capacity-constrained relaxation.}
For fixed continuous capacities $\widetilde{\mathbf c}=(\widetilde c_1,\ldots,\widetilde c_M)^\top$, the entropy-regularized assignment problem is
\begin{equation}
\begin{aligned}
 \max_{\mathbf P\geq0}\quad&\langle\mathbf S,\mathbf P\rangle+\varepsilon\mathcal H(\mathbf P)\\
 \text{subject to}\quad&\mathbf P\mathbf1_M=\mathbf1_N,\qquad
 \mathbf P^\top\mathbf1_N=\widetilde{\mathbf c},
\end{aligned}
 \label{eq:transport-relaxation}
\end{equation}
where $\mathbf S=(S_{ij})\in\mathbb R^{N\times M}$, $\mathbf P\in\mathbb R_+^{N\times M}$, $\mathbf1_k$ is the length-$k$ all-ones vector, and $\mathcal H(\mathbf P)=-\sum_{ij}P_{ij}\log P_{ij}$ with $0\log0=0$. The parameter $\varepsilon>0$ controls entropy regularization. Sinkhorn scaling alternately balances rows and columns of the exponential affinity kernel~\citep{cuturi2013sinkhorn}. Training uses a finite-iteration approximation $\widehat{\mathbf P}$ to these marginals. The confidence-ordered discrete assignment is a separate capacity-preserving procedure, not an exact optimizer of Equation~(\ref{eq:transport-relaxation}).

\Needspace{4\baselineskip}
\paragraph{Discrete pooling with surrogate gradients.}
Write $\mathbf W^{\mathrm{hard}}$ for the pooling matrix supported on the discrete groups. Column normalization of $\widehat{\mathbf P}$ gives a differentiable relaxation in a straight-through-style surrogate~\citep{bengio2013estimating}:
\begin{equation}
 W^{\mathrm{soft}}_{ij}=\frac{\widehat P_{ij}}{\sum_{k=1}^{N}\widehat P_{kj}},
 \qquad \mathbf W^{\mathrm{train}}=\mathbf W^{\mathrm{hard}}+\mathbf W^{\mathrm{soft}}
                 -\operatorname{sg}(\mathbf W^{\mathrm{soft}}).
\label{eq:pooling-surrogate}
\end{equation}
The stop-gradient operator $\operatorname{sg}$ preserves its argument's value but contributes zero gradient during differentiation. Thus the pooling weights equal $\mathbf W^{\mathrm{hard}}$, while their gradient includes both the within-group softmax term and the relaxation's contribution to capacity and cross-group affinity learning. The discrete group assignments are held fixed during differentiation; recall always accesses the original Fine members indexed by C2F.

\subsubsection{Group-Capacity Sensitivity}
\label{app:capacity-analysis}
\paragraph{Sensitivity under a fixed total.}
Assume $U>1$, $M<N<MU$, and finite scores and priors. Define $d_j=(U-1)\operatorname{sigmoid}'(b_j+a_j+\mu)>0$, $\mathbf d=(d_1,\ldots,d_M)^\top$, and $D=\sum_jd_j$. The unique continuous allocation has Jacobian
\begin{equation}
 J_{jk}=\frac{\partial\widetilde c_j}{\partial a_k}
 =d_j\left(\mathbf1_{\{j=k\}}-\frac{d_k}{D}\right),
 \qquad \mathbf J=\operatorname{diag}(\mathbf d)-\frac{\mathbf d\mathbf d^\top}{D}.
 \label{eq:capacity-sensitivity}
\end{equation}
The matrix $\mathbf J$ is symmetric positive semidefinite, has nullspace $\operatorname{span}\{\mathbf1_M\}$, and has rank $M-1$. Its column sums vanish. For $M>1$, increasing one score alone increases that group's capacity and decreases every other group's capacity.

\paragraph{Proof.}
The capacity sum is strictly increasing in $\mu$, with limits $M$ and $MU$, establishing the unique finite offset. Differentiating the sum constraint gives $0=d_k+D\,\partial\mu/\partial a_k$, which yields Equation~(\ref{eq:capacity-sensitivity}). For any $\mathbf v\in\mathbb R^M$, let $\bar v_d=D^{-1}\sum_jd_jv_j$. Then
\begin{equation}
 \mathbf v^\top\mathbf J\mathbf v
 =\sum_jd_j(v_j-\bar v_d)^2\geq0.
\end{equation}
Since every $d_j>0$, equality holds exactly when all coordinates of $\mathbf v$ are equal. This proves the nullspace and rank statements. A common score shift is absorbed by $\mu$, leaving the allocation unchanged.

\paragraph{Relative marginal-loss signal.}
For a differentiable relaxed loss, let $\xi_j=\partial\widetilde{\mathcal L}/\partial\widetilde c_j$. Holding other computational paths fixed, the chain rule gives
\begin{equation}
 \left.\frac{\partial\widetilde{\mathcal L}}{\partial a_k}\right|_{\mathrm{capacity}}
 =d_k(\xi_k-\bar\xi_d),\qquad
 \bar\xi_d=\frac{\sum_jd_j\xi_j}{D}.
 \label{eq:capacity-loss-gradient}
\end{equation}
The capacity predictor receives each region's marginal loss relative to a weighted regional mean. These derivatives characterize the continuous allocation; the shared predictor couples score updates, and discrete grouping is trained through the surrogate in Equation~(\ref{eq:pooling-surrogate}).

\subsection{TRACE: Temporal Routing and Evidence Access}
\subsubsection{Parameterization and History}
\label{app:history-analysis}
\paragraph{Branch parameterization.}
For Equation~(\ref{eq:ssm}), each branch has a learned decay parameter $\boldsymbol\theta_m\in\mathbb R^{d_s}$. The input-dependent decay, shared step-size and write maps, and branch mixture are
\begin{equation}
\begin{aligned}
 \boldsymbol\rho_t^{(m)}&=\exp\!\left[-\operatorname{softplus}(\boldsymbol\theta_m)
                   \odot\boldsymbol\Delta(\mathbf u_t)\right],\\
 \boldsymbol\Delta(\mathbf u_t)&=\operatorname{softplus}(\mathbf W_\Delta\mathbf u_t+\mathbf b_\Delta),\qquad
 f(\mathbf u_t)=\mathbf W_w\mathbf u_t,\\
 \boldsymbol\beta_t&=\operatorname{softmax}(\mathbf W_\beta\mathbf u_t).
\end{aligned}
\label{eq:branch-parameters}
\end{equation}
Here $\operatorname{softplus}(x)=\log(1+e^x)$ acts elementwise. Softmax acts over branches, giving positive weights that sum to one. Appendix~\ref{app:training-architecture} gives the concrete architecture and timescale initialization.

\paragraph{How past states contribute.}
With zero initial state, unrolling a branch gives
\begin{equation}
 \mathbf s_t^{(m)}=\sum_{\tau=1}^{t}
 \underbrace{\left[(\mathbf1-\boldsymbol\rho_\tau^{(m)})\odot
 \prod_{k=\tau+1}^{t}\boldsymbol\rho_k^{(m)}\right]}_{\boldsymbol\alpha_{t,\tau}^{(m)}}
 \odot f(\mathbf u_\tau).
 \label{eq:ssm-history}
\end{equation}
Products are elementwise and empty products equal one. The formula follows by induction: multiplication by $\boldsymbol\rho_t^{(m)}$ extends all preceding retention factors, and the new write adds the $\tau=t$ term. For finite parameters, $\mathbf0<\boldsymbol\rho_t^{(m)}<\mathbf1$, and
\begin{equation}
 \sum_{\tau=1}^{t}\boldsymbol\alpha_{t,\tau}^{(m)}
 =\mathbf1-\prod_{k=1}^{t}\boldsymbol\rho_k^{(m)}\leq\mathbf1.
\end{equation}
Each state coordinate is consequently a convex combination of historical write coordinates and its zero initial value. If $L_t=\max_{\tau\leq t}\|f(\mathbf u_\tau)\|_\infty$, then $\|\mathbf s_t^{(m)}\|_\infty\leq L_t$ and $\|\mathbf z_t\|_\infty\leq L_t$. The retention factors expose how each branch weights past writes, while the learned mixture adapts their contribution to the current decision.

\subsubsection{Evidence Policy and Decoding Algorithm}
\label{app:cache-interface}
\paragraph{Proposing evidence.}
Projections $q$ and $k$ map the temporal state and Coarse descriptors to dimension $d_r$. A learned scalar function $e$ sets the selection boundary:
\begin{equation}
\begin{aligned}
 r_{t,j}&=\frac{q(\mathbf z_t)^\top k(\mathbf x_j^{\mathrm C})}{\sqrt{d_r}},
 &p_{t,j}&=\operatorname{sigmoid}\!\left(r_{t,j}-e(\mathbf z_t)\right),\\
 \widehat{\mathcal A}_t&=\{j\in[M]:p_{t,j}>1/2\}.
\end{aligned}
\label{eq:candidate}
\end{equation}

\paragraph{Comparing retained and proposed evidence.}
The gate in Equation~(\ref{eq:commit}) summarizes the current set and the proposed memberships in the same region-key space. With $\mathbf k_j=k(\mathbf x_j^{\mathrm C})$, its evidence inputs are
\begin{equation}
 \bar{\mathbf k}_t^{\mathrm{act}}
 =\frac{\sum_{j\in\mathcal A_{t-1}}\mathbf k_j}{\max(1,|\mathcal A_{t-1}|)},
 \qquad \bar{\mathbf k}_t^{\mathrm{prop}}
 =\frac{\sum_{j=1}^{M}p_{t,j}\mathbf k_j}
 {\epsilon_g+\sum_{j=1}^{M}p_{t,j}}.
\label{eq:gate-summaries}
\end{equation}
The empty-set summary is zero and $\epsilon_g>0$ stabilizes normalization. Together with $\mathbf z_t$, these summaries let the gate condition retention on both the available evidence and the newly predicted demand. At inference, $g_t$ is the more probable HOLD or UPDATE action under $\pi_\psi$.

\paragraph{Initial visual context.}
Before generating the first response token, ViMoD switches the active visual context from full visual KV to Coarse KV and the initially selected Fine KV. The full Fine KV bank remains stored for later recall.

\begin{algorithm}[htbp]
\caption{ViMoD inference}
\label{alg:fomer}
\begin{algorithmic}[1]
\REQUIRE Image $I$, prompt $x$; parameters $\theta_{\mathrm{Frozen}},\phi,\psi$
\ENSURE Generated response
\STATE Extract $\mathbf X^{\mathrm F}$; prefill and cache $\mathrm{KV}^{\mathrm F},\mathrm{KV}^{\mathrm T}$
\STATE $(\mathbf X^{\mathrm C},\mathrm{KV}^{\mathrm C},\Pi)\gets\operatorname{DART}_{\phi}(\mathbf X^{\mathrm F},\mathrm{KV}^{\mathrm F})$ (Equations~\ref{eq:capacity}--\ref{eq:coarse-kv})
\STATE $\mathcal A_0\gets\varnothing$; $\mathbf s_0^{(m)}\gets\mathbf0$ for all $m$
\FOR{each routing step $t$}
  \STATE Compute $\mathbf u_t$; update $\{\mathbf s_t^{(m)}\}_m,\mathbf z_t$ (Equation~\ref{eq:ssm})
  \STATE Compute proposal $\widehat{\mathcal A}_t$ and gate $g_t$ (Equations~\ref{eq:candidate} and~\ref{eq:commit})
  \STATE $\mathcal A_t\gets\mathcal A_{t-1}$ if $g_t=\textsc{Hold}$, otherwise $\widehat{\mathcal A}_t$
  \STATE $\mathcal F_t\gets\bigcup_{j\in\mathcal A_t}\Pi(j)$
  \STATE Decode with Joint-KV until the next routing step or termination (Equation~\ref{eq:joint-kv})
\ENDFOR
\end{algorithmic}
\end{algorithm}

\subsection{Training Objectives}
\label{app:training-objectives}
\subsubsection{Compact Memory Distillation}
DART matches full-visual teacher predictions on task-verified responses under identical teacher-forced prefixes:
\begin{equation}
 \mathcal L_{\mathrm{DART}}=
 \mathcal L_{\mathrm{CE}}+
 \lambda_{\mathrm{KD}}\mathcal L_{\mathrm{KL}}.
 \label{eq:distillation}
\end{equation}
The terms supervise response tokens by cross-entropy and $D_{\mathrm{KL}}(p_{\mathrm{full}}\,\|\,p_{\theta_{\mathrm{Frozen}},\phi})$, where $p_{\theta_{\mathrm{Frozen}},\phi}$ uses Coarse visual memory. Gradients train capacities, affinities, and Value aggregation through the discrete pooling surrogate in Equation~(\ref{eq:pooling-surrogate}). Appendix~\ref{app:training-recipe} gives the distillation configuration.

\subsubsection{Supervised Set and Gate Learning}
For segment $s$, $\mathcal E_s$ contains Fine cells with positive-area overlap with the teacher evidence boxes. One segment may supervise several routing decisions.

Set training updates TRACE parameters $\psi$ except the gate. Let $\mathcal P_t=\mathcal A_t^\star$ and $\mathcal Q_t=[M]\setminus\mathcal P_t$ be the positive and negative region sets. The class-balanced loss at routing step $t$ is
\begin{equation}
 \ell_{\mathrm{set},t}
 =-\frac1{\max(1,|\mathcal P_t|)}\sum_{j\in\mathcal P_t}\log p_{t,j}
  -\frac1{\max(1,|\mathcal Q_t|)}\sum_{j\in\mathcal Q_t}\log(1-p_{t,j}).
\end{equation}
Region BCE is averaged separately over positive and negative regions and summed; an absent class contributes zero. Joint training adds class-balanced gate cross-entropy. The gate target is UPDATE when consecutive annotated sets differ and HOLD otherwise. Sec.~\ref{sec:training} defines the causal supervision targets, and Appendix~\ref{app:training-recipe} specifies the trainable components of each stage.

\subsubsection{Policy Optimization and Online Supervision}
\paragraph{Sampling and trajectory feedback.}
Policy training samples $g_t$ from its categorical distribution. On UPDATE, region memberships are independently sampled as $m_{t,j}\sim\operatorname{Bern}(p_{t,j})$, replacing the previous set; on HOLD, the active set persists. For $G=16$ rollouts per question and correctness $Y_i\in\{0,1\}$, RLOO~\citep{ahmadian2024back} gives the leave-one-out correctness advantage
\begin{equation}
 A_i^{\mathrm{acc}}=2Y_i-\frac1{G-1}\sum_{j\ne i}2Y_j.
\end{equation}
For trajectory $i$, let $N_i$ be its original Fine-token count, $T_i>0$ its number of generation forward passes, and $t_i(n)$ the latest routing decision before pass $n$. Its mean Fine occupancy is
\begin{equation}
 C_i=\frac{1}{N_iT_i}
       \sum_{n=1}^{T_i}|\mathcal F_{i,t_i(n)}|.
 \label{eq:access-cost}
\end{equation}
Let $\mathcal I_+=\{i:Y_i=1\}$, with costs $C_i$ from Equation~(\ref{eq:access-cost}). When $|\mathcal I_+|\geq2$, define $\bar C_+=|\mathcal I_+|^{-1}\sum_{j\in\mathcal I_+}C_j$ and $A_i^C=\bar C_+-C_i$ for $i\in\mathcal I_+$; all other cost advantages are zero. These costs measure Fine residency, including HOLD intervals, and exclude persistent Coarse tokens. Cost advantages are not divided by their standard deviation.

Define the trajectory log-probabilities
\begin{equation}
\begin{aligned}
 \ell_i^{\mathrm{region}}&=\sum_{t:g_{i,t}=\textsc{Update}}\sum_j
 \Big[m_{i,t,j}\log p_{i,t,j}+(1-m_{i,t,j})\log(1-p_{i,t,j})\Big],\\
 \ell_i^{\mathrm{all}}&=\sum_t\log\pi_\psi(g_{i,t}\mid\cdot)+\ell_i^{\mathrm{region}}.
\end{aligned}
\end{equation}
The symbol $\cdot$ includes the temporal state, previous active set, proposal probabilities, and Coarse descriptors, together with the update-specific exploration bias $b$ (see Appendix~\ref{app:training-recipe} for the training recipe). The per-question policy loss is
\begin{equation}
 \mathcal L_{\mathrm{policy}}=-\frac1G\sum_{i=1}^G
 \left[A_i^{\mathrm{acc}}\ell_i^{\mathrm{all}}+0.5A_i^C\ell_i^{\mathrm{region}}\right].
 \label{eq:policy}
\end{equation}
The batch policy loss averages over questions, with advantages held fixed during differentiation. Correctness weights both gate and region actions; cost directly weights only region log-probabilities, although shared representations can also change gate behavior. Success-conditioned cost feedback imposes no hard accuracy or computation constraint. Gradients propagate through recurrent router states, while backbone hidden states and observed trajectory histories are fixed observations.

\paragraph{Online teacher guidance.}
The teacher sees the image, question, and generated prefix, but neither the reference answer nor future student text. Its boxes specify a complete working set for approximately the next 16 student tokens. After the mapping in Appendix~\ref{app:training-recipe}, let $\mathcal V_{i,t}$ be the valid regions and $y_{i,t,j}\in\{0,1\}$ their labels. For annotations containing both positive and negative regions, we use binary cross-entropy with $p_{i,t,j}=\operatorname{sigmoid}(r_{i,t,j}-e(\mathbf z_{i,t}))$:
\begin{equation}
 \ell_{\mathrm{Teacher},i,t}
 =-\frac1{|\mathcal V_{i,t}|}\sum_{j\in\mathcal V_{i,t}}
 \Big[y_{i,t,j}\log p_{i,t,j}+(1-y_{i,t,j})\log(1-p_{i,t,j})\Big].
\end{equation}
We average over valid regions and then valid annotation positions within each trajectory, skipping empty, all-positive, and uncertain annotations. Thus an empty box list supplies no BCE gradient. Unlike SFT, this loss does not balance the classes; it supervises region scores, the selection threshold, and shared representations, and can therefore change selected-set size.

\paragraph{Auxiliary update signal.}
When all $G$ rollouts for a question fail and none contains an autonomous UPDATE, we apply the trajectory-level auxiliary term
\begin{equation}
 \mathcal L_{\mathrm{call},i}=-\log\!\left[1-\prod_t(1-q_{i,t})\right],
\end{equation}
where $q_{i,t}$ is the UPDATE probability at a state observed along the sampled trajectory, conditioned on the same exploration bias. This surrogate encourages updates at those states; it is not the exact probability of an update under closed-loop generation. The complete objective is
\begin{equation}
 \mathcal L_{\mathrm{RL}}=
 \mathcal L_{\mathrm{policy}}+0.3\mathcal L_{\mathrm{Teacher}}+0.1\mathcal L_{\mathrm{call}}.
 \label{eq:training-router}
\end{equation}
Appendix~\ref{app:training-recipe} gives the sampling, annotation, and optimization settings.

%% file: appendices/training_details.tex
\section{Training Details}
\label{app:training-details}
\label{app:training-recipe}
\input{appendices/training_data}
\input{appendices/training_configuration}

%% file: appendices/training_data.tex
\subsection{Training Data Sources}
\label{app:training-data}

\paragraph{DART.} The 57,420 trajectories cover mathematics and science (Euclid30K~\citep{lian2026euclid}, Geo170K~\citep{gao2025g}, M3CoT~\citep{chen2024m3cot}, MAVIS~\citep{zhang2025mavis}, ScienceQA~\citep{lu2022learn}), natural-image understanding (GQA~\citep{hudson2019gqa}, Visual7W~\citep{zhu2016visual7w}), documents, charts, and tables (PixMo~\citep{deitke2025molmo}, PlotQA~\citep{methani2020plotqa}), OCR (MathWriting~\citep{gervais2025mathwriting}, TextVQA~\citep{singh2019towards}, UniMER~\citep{wang2024unimernet}, ReCTS~\citep{zhang2019icdar}, MLT2019~\citep{nayef2019icdar2019}), and localization (RefCOCO~\citep{yu2016modeling}, Visual Genome~\citep{krishna2017visual}, PixMo points~\citep{deitke2025molmo}).

\begin{figure}[htbp]
\centering
\includegraphics[trim=10.5bp 5bp 11.5bp 2.5bp,clip,width=0.96\linewidth]{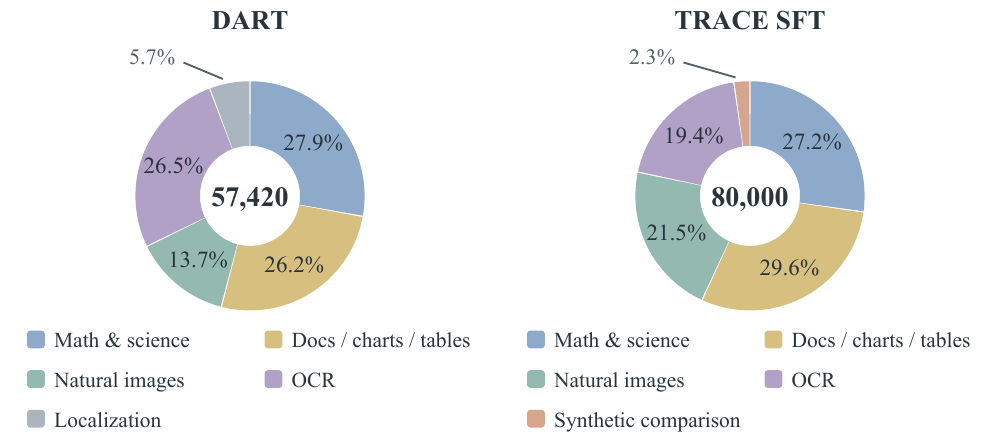}
\figcaptionsep
\caption{Training-data composition by task type. The two DART compressors share one dataset, as do the two TRACE SFT stages; the RL pool is a subset of TRACE SFT and is not counted separately.}
\label{fig:training-data-composition}
\end{figure}

\paragraph{TRACE.} The 80,000 SFT examples span mathematics and science (Euclid30K~\citep{lian2026euclid}, Geo170K~\citep{gao2025g}, M3CoT~\citep{chen2024m3cot}, MAVIS~\citep{zhang2025mavis}, OpenMMReasoner~\citep{zhang2026openmmreasoner}, ScienceQA~\citep{lu2022learn}), documents, charts, and tables (PixMo~Documents~\citep{deitke2025molmo,yang2025scaling}, PlotQA~\citep{methani2020plotqa}), natural images (GQA~\citep{hudson2019gqa}, Visual7W~\citep{zhu2016visual7w}), OCR (MathWriting~\citep{gervais2025mathwriting}, TextVQA~\citep{singh2019towards}, UniMER~\citep{wang2024unimernet}), and synthetic visual comparisons (SuperCLEVR Compare~\citep{li2023super}). The 4,800 RL questions are a task-balanced subset of this SFT set. All training data are disjoint from every evaluation set. Fig.~\ref{fig:training-data-composition} compares the task-type proportions; the supplementary dataset breakdown provides per-source counts.

%% file: appendices/training_configuration.tex
\subsection{Architecture and Optimization Setup}
\label{app:training-architecture}

\paragraph{Common setup.} Training uses NVIDIA H200 GPUs. The frozen \modelname{Qwen3-VL-4B-Instruct}~\citep{bai2025qwen3} backbone has about 4.44 billion parameters; DART and TRACE add approximately 1.185 million and 1.236 million parameters, respectively, totaling 0.0546\% of the backbone. The annotation teacher is excluded from deployment counts. Tab.~\ref{tab:training-hparams} summarizes the schedule and trainable components; Tab.~\ref{tab:trace-config} specifies TRACE's state-space architecture. The backbone remains frozen throughout, and DART is frozen during all TRACE training. All stages use AdamW~\citep{loshchilov2017decoupled}; DART and SFT use zero weight decay. Batch sizes are global across devices and gradient accumulation.

\begin{table}[htbp]
\caption{Training schedule and updated components. Dataset sizes count DART response trajectories, SFT examples, and RL questions. The two SFT stages share one 80,000-example set. DART is trained independently at each ratio; all TRACE stages use 9:1 compression.}
\label{tab:training-hparams}
\centering\fontsize{9}{11}\selectfont
\setlength{\tabcolsep}{3pt}
\renewcommand{\arraystretch}{1.45}
\begin{tabular}{@{}lllll@{}}
\toprule
\textbf{Setting} & \textbf{DART} & \textbf{TRACE SFT: Set} & \textbf{TRACE SFT: Joint} & \textbf{TRACE RL} \\
\midrule
Dataset size & 57,420 & 80,000 & 80,000 & 4,800 \\
Training duration & 1,200 steps & 3 epochs & 3 epochs & 50 steps \\
Global batch size & 192 & 64 & 64 & 96 \\
Learning rate & $10^{-5}$ & $10^{-4}$ & $10^{-4}$ & $3\times10^{-5}$ \\
\midrule
Updated components & DART & TRACE except gate & Layer fusion, adapter, SSM, gate & Entire TRACE \\
\bottomrule
\end{tabular}
\end{table}

\begin{table}[htbp]
\caption{TRACE architecture and routing configuration.}
\label{tab:trace-config}
\centering\small
\setlength{\tabcolsep}{4pt}
\renewcommand{\arraystretch}{1.45}
\begin{tabular}{@{}>{\raggedright\arraybackslash}p{0.40\linewidth}>{\raggedright\arraybackslash}p{0.54\linewidth}@{}}
\toprule
\textbf{Setting} & \textbf{Value} \\
\midrule
Input hidden dimension & 2,560 \\
Backbone hidden-state indices & $27,31,35$ (backbone indexing) \\
SSM state dimension & 256 \\
Region key/query dimension & 128 \\
Initial timescales & 1, 2, 4, 8 routing steps \\
Routing interval & 16 generated tokens \\
\bottomrule
\end{tabular}
\end{table}

\paragraph{Architecture details.} The two DART configurations use the same parameter dimensions, with group-size ranges $[1,8]$ for 4:1 and $[1,18]$ for 9:1. Assignment projections have dimension 128 and use eight balancing iterations. For each KV head, the Value encoder uses a residual transformation comprising layer normalization~\citep{ba2016layer} and two linear layers separated by a Gaussian error linear unit (GELU) activation~\citep{hendrycks2016gaussian}. Its parameters are shared across decoder layers; the head dimension is 128 and the intermediate dimension is 256. TRACE's region keys and queries use root-mean-square normalization~\citep{zhang2019root} followed by a linear projection without a bias term; the selection threshold is an affine function of the state.

\subsection{Stagewise Training Procedure}
\subsubsection{DART Distillation} The 4:1 and 9:1 compressors start from $2\times2$ and $3\times3$ regular-grid mean pooling, respectively, and use identical data and optimization settings. The frozen backbone generates responses with full visual context. We retain trajectories that pass task-specific checks of answers, recognized text, or bounding boxes. Prompts request brief evidence for visual question answering, allow reasoning for mathematics and science, and request text or normalized boxes for OCR and localization.

Under teacher forcing, the full-context backbone supplies token distributions, while the student re-encodes the text prefix with attention to the compressed Coarse visual keys and values. Equation~(\ref{eq:distillation}) uses $\lambda_{\mathrm{KD}}=2$ at temperature 1. We supervise at most the first 64 answer tokens, including the first token. DART training uses no warmup and a gradient-clipping threshold of 1.

\subsubsection{TRACE Supervised Training} \modelname{Qwen3.8-27B}~\citep{qwen38} annotates the 80,000-example SFT set using greedy decoding and the offline template in Appendix~\ref{app:teacher-prompts} (Prompt~1). Tab.~\ref{tab:teacher-config} distinguishes offline annotation from online supervision. We check answer correctness, output format, and coordinates before mapping boxes to compressor regions. Processing the teacher response causally through the frozen student provides hidden states, region features, and working-set history, with routing targets aligned every 16 tokens.

The Set stage uses class-balanced set binary cross-entropy (BCE). The Joint stage uses set BCE plus class-balanced gate cross-entropy (loss details in Appendix~\ref{app:training-objectives}). Its region key/query projections, associated normalization parameters, and selection-threshold parameters remain frozen; Tab.~\ref{tab:training-hparams} lists the components updated in each stage.

\begin{table}[htbp]
\caption{Offline and online evidence annotation. Both use \modelname{Qwen3.8-27B} (\codeparam{temperature=0}, optional reasoning disabled) and image-relative box coordinates in $[0,1000]$. Complete templates appear in Appendix~\ref{app:teacher-prompts}.}
\label{tab:teacher-config}
\centering\small
\setlength{\tabcolsep}{4pt}
\renewcommand{\arraystretch}{1.15}
\begin{tabular}{@{}>{\raggedright\arraybackslash}p{0.16\linewidth}>{\raggedright\arraybackslash}p{0.42\linewidth}>{\raggedright\arraybackslash}p{0.37\linewidth}@{}}
\toprule
\textbf{Setting} & \textbf{Offline SFT} & \textbf{Online RL} \\
\midrule
Inputs & Image and question & Image, question, and student prefix \\
Prediction target & Reasoning steps, evidence boxes for each upcoming step, and final answer & Evidence boxes for approximately the next 16 student tokens \\
Output format & XML with JSON evidence fields & Schema-constrained JSON \\
\bottomrule
\end{tabular}
\end{table}

\subsubsection{TRACE Reinforcement Learning} RL follows 9:1 joint supervised training with a newly initialized AdamW optimizer. The 4,800-question pool is curated from SFT using correctness-based filtering and task-type balancing. Screening samples four trajectories per question from the fixed 9:1 SFT model and retains groups with both successes and failures, or groups in which every trajectory fails without an autonomous UPDATE. Training uses the pool without replacement in 50 batches of 96 questions, balancing task types within each batch. Each question yields $G=16$ newly sampled trajectories; each batch produces one optimizer update, without a critic or PPO clipping. RL rollouts use \codeparam{max\_new\_tokens=4096}.

\paragraph{Online evidence supervision.} The online teacher uses Prompt~2 in Appendix~\ref{app:teacher-prompts}, with \codeparam{max\_new\_tokens=512} and annotations at up to 24 routing decisions per trajectory. It receives neither reference answers nor future student text. Its boxes specify a complete proposed working set.

The training objectives are defined in Appendix~\ref{app:training-objectives}.

\paragraph{Exploration.} At each optimization step, we sample one gate bias $b\sim\mathcal U(-2.5,2.5)$, shared by the rollout batch and held fixed when evaluating the sampled actions' probabilities for the update. This bias conditions the policy during exploration and is not an additional loss.

%% file: appendices/experimental_details.tex
\section{Evaluation Protocols and Efficiency Accounting}
\label{app:experimental-details}

\subsection{Benchmarks and Metrics}
\label{app:benchmark-provenance}
The reasoning suite comprises We-Math~\citep{qiao2025we}, DynaMath~\citep{zou2025dynamath}, MathVerse~\citep{zhang2024mathverse}, MathVista~\citep{lu2024mathvista}, MathVision~\citep{wang2024measuring}, LogicVista~\citep{xiao2024logicvista}, VisualPuzzles~\citep{song2025visualpuzzles}, and MMMU-Pro~\citep{yue2025mmmu}. The general visual understanding and document--chart QA suite comprises GQA~\citep{hudson2019gqa}, POPE~\citep{li2023evaluating}, V$^*$~\citep{wu2024v}, MME~\citep{fu2026mme}, MMBench~\citep{liu2024mmbench}, DocVQA~\citep{mathew2021docvqa}, OCRBench v2~\citep{fu2026ocrbench}, and ChartQA~\citep{masry2022chartqa}. Explicit CoT prompts are marked in Tab.~\ref{tab:main}.

We follow the official benchmark scoring rules as implemented in VLMEvalKit~\citep{duan2024vlmevalkit}. We use GQA~\citep{hudson2019gqa} TestDev Balanced, MMBench~\citep{liu2024mmbench} English DEV v1.0, and the ChartQA~\citep{masry2022chartqa} test split.

\paragraph{Generation settings and reference scores.}
\label{app:generation-protocol}
All compared methods, including Full, use greedy decoding with \codeparam{presence\_penalty=0}. Output limits vary across tasks and are shared by methods within each comparison. The main MMMU-Pro benchmark uses \codeparam{max\_new\_tokens=4096}; the separate efficiency evaluation in Fig.~\ref{fig:efficiency-tradeoff} uses 8,192 (see Appendix~\ref{app:efficiency-measurement}).

The \modelname{Qwen3-VL-4B-Instruct} model card recommends sampling for visual tasks with temperature 0.7, top-$p$ 0.8, top-$k$ 20, repetition penalty 1.0, presence penalty 1.5, and an output limit of 16,384 tokens.\footnote{\url{https://huggingface.co/Qwen/Qwen3-VL-4B-Instruct}} Our use of greedy decoding and zero presence penalty differs from these recommendations across the evaluation suite, limiting direct comparability with externally reported backbone scores.

As one example, our Full baseline obtains 30.92\% on MMMU-Pro under our evaluation protocol; the backbone's technical report lists 53.2\%~\citep{bai2025qwen3}. Repetition and truncation may affect results, but we have not isolated their contribution to this gap or measured the accuracy and response-length effects of alternative settings across benchmarks. All normalized results use our measured Full baseline under the same protocol as the compared methods.

\subsection{Visual-Budget Calibration}
\label{app:budget-calibration}
For ViMoD, average visual occupancy is calibrated separately for GQA, MME, MMBench, and ChartQA and for the remaining evaluation suite.
To reach a target average total visual occupancy $\nu$, deployment adds a scalar offset $\delta_\nu$ to TRACE's learned boundary:
\begin{equation}
 p_{t,j}^{(\nu)}=\operatorname{sigmoid}\!\left(r_{t,j}-e(\mathbf z_t)-\delta_\nu\right),
 \qquad \widehat{\mathcal A}_t^{(\nu)}=\{j:p_{t,j}^{(\nu)}>\tfrac12\}.
\end{equation}
For a nonempty trajectory of response-generation forward passes, Fine occupancy $C_i$ is defined in Equation~(\ref{eq:access-cost}); total visual occupancy is
\begin{equation}
 R_i=\frac{M_i}{N_i}+C_i
 =\frac1{N_iT_i}
 \sum_{n=1}^{T_i}\left(M_i+|\mathcal F_{i,t_i(n)}|\right).
\end{equation}
Since $0\leq C_i\leq1$, total occupancy lies in $[M_i/N_i,\,1+M_i/N_i]$; the persistent Coarse fraction sets a lower bound on feasible usage. Calibration adjusts $\delta_\nu$ on a moderate sample until the measured average total occupancy $\widehat R_{\mathrm{cal}}(\delta_\nu)$ satisfies
\begin{equation}
 \bigl|\widehat R_{\mathrm{cal}}(\delta_\nu)-\nu\bigr|\leq\epsilon_{\mathrm{cal}},
\end{equation}
where $\epsilon_{\mathrm{cal}}$ is the calibration tolerance. The offset is then fixed for evaluation. The calibrated probabilities replace $p_{t,j}$ in both candidate selection and the proposal summary in Equation~(\ref{eq:gate-summaries}). Holding scores fixed, increasing the offset can only remove candidate regions; closed-loop occupancy also depends on gate decisions, subsequent states, and generated length, so it need not be monotone.

\input{appendices/cost_accounting}

\subsection{Diagnostic Settings}
\label{app:diagnostic-settings}
\paragraph{Visual-demand variation.}
For the diagnostics in Figs.~\ref{fig:dynamic-demand} and~\ref{fig:answer-retention}(a), we use responses generated by the full-context \modelname{Qwen3-VL-4B-Instruct} backbone. Given a response of $T$ tokens, we partition it into $N_{\mathrm{win}}$ contiguous, approximately equal-length windows with boundaries $b_k=\lfloor kT/N_{\mathrm{win}}\rfloor$ and $b_0=0$. Let $\mathbf h_i$ be the embedding of visual token $i$ at the language-model input, and let $\mathcal W_k=\{b_{k-1}+1,\ldots,b_k\}$. Along the recorded generation trajectory, the gradient-times-input attribution~\citep{ancona2017towards} of visual token $i$ to window $k$ and its within-window normalization are
\begin{equation}
 R_{k,i}=
 \left\|\mathbf h_i\odot\nabla_{\mathbf h_i}
 \sum_{t\in\mathcal W_k}\log p_\theta(y_t\mid I,x,y_{<t})\right\|_2,
 \qquad
 P_{k,i}=\frac{R_{k,i}}{\sum_j R_{k,j}}.
\end{equation}
The gradient is taken with respect to the visual input embedding, with backbone parameters fixed; $P_{k,i}$ is an attribution score, not an attention weight. For $N_{\mathrm{vis}}$ visual tokens, let $m=\max(1,\lceil0.1N_{\mathrm{vis}}\rceil)$ and let $\mathcal S_k$ contain the $m$ tokens with the largest $P_{k,i}$. We define visual-demand variation as their mean turnover between adjacent windows:
\begin{equation}
 V=\frac{1}{N_{\mathrm{win}}-1}\sum_{k=2}^{N_{\mathrm{win}}}
 \left(1-\frac{|\mathcal S_k\cap\mathcal S_{k-1}|}{m}\right).
\end{equation}
The reported diagnostic uses $N_{\mathrm{win}}=16$; Fig.~\ref{fig:dynamic-demand} displays selected windows, whereas $V$ uses all of them. Fig.~\ref{fig:answer-retention}(a) groups MathVerse examples by $V$, computed from their full-context responses, and plots $100V$ on the horizontal axis. Each displayed heatmap is additionally scaled by its own maximum for color rendering; this display scaling is not used to compute $V$.

\paragraph{Generation-cost diagnostic.}
The FastV~\citep{chen2024image} diagnostic in Fig.~\ref{fig:reasoning-cost} uses 300 MathVerse~\citep{zhang2024mathverse} testmini examples with \codeparam{max\_new\_tokens=8192}. All retention settings use the same example set and generation cap. Early entropy is the mean next-token entropy over the first eighth of each response. Text similarity is the mean cosine similarity between embeddings of neighboring 128-token windows with stride 64, computed for responses with at least 256 tokens. Amortized decoder FLOPs per generated token are computed as total decoder FLOPs divided by output length.

\paragraph{Annotation-guided evidence-access comparison.}
For Fig.~\ref{fig:access-tradeoff}(b), \modelname{Qwen3.8} annotates the visual regions required at each reasoning stage on MMMU-Pro. Dynamic access retains persistent visual context alongside the selected stage-specific regions. The vertical axis reports accuracy as a percentage of the uncompressed Full model's accuracy. The budgets marked at 70\% are obtained by linear interpolation between neighboring plotted points.

%% file: appendices/cost_accounting.tex
\subsection{Cost Model and FLOP Accounting}
\label{app:cost-accounting}

\subsubsection{Visual Occupancy and Generation Cost}
\label{app:cost-analysis}
\label{sec:cost-analysis}
For a response of $T$ generation forward passes, the visual-token budget measures average active occupancy relative to the $N$ original visual tokens:
\begin{equation}
 R=\frac{1}{NT}\sum_{n=1}^{T}\left(M+|\mathcal F_{t(n)}|\right).
 \label{eq:visual-budget}
\end{equation}
This counts persistent Coarse tokens and active Fine tokens on every pass, including HOLD intervals. The recorded circuit case illustrates Fine access, Coarse-only reasoning, and renewed access (Fig.~\ref{fig:qualitative-circuit}).

To relate the occupancy objective in Equation~(\ref{eq:access-cost}) to visual-attention work, assume cost $\kappa>0$ per active visual KV entry per forward pass. With total occupancy $R$ from Equation~(\ref{eq:visual-budget}), this work is $\kappa\sum_{n=1}^{T}(M+|\mathcal F_{t(n)}|)=\kappa NTR$. Thus dynamic and fixed contexts with equal $R$ incur equal modeled visual-attention work for the same image and generation length. Let $B_{\mathrm{init}}$ be initialization cost beyond the uncompressed baseline and $B_{\mathrm{route}}(T)$ the routing overhead. At a common horizon of $T$ forward passes with unchanged nonvisual computation, modeled savings are positive when
\begin{equation}
 \underbrace{B_{\mathrm{init}}+B_{\mathrm{route}}(T)}_{\text{additional overhead}}
 <\underbrace{\kappa\sum_{n=1}^{T}(N-M-|\mathcal F_{t(n)}|)}_{\text{saved visual-attention work}}.
 \label{eq:break-even}
\end{equation}
For trajectories of different lengths, let $T_F$ and $T_M$ denote the numbers of forward passes for the uncompressed baseline and ViMoD, and $Q_F(T_F)$ and $Q_M(T_M)$ their nonvisual response computation, including text attention. The total-cost difference becomes
\begin{equation}
\begin{split}
 C_{\mathrm{ViMoD}}-C_{\mathrm{Full}}={}&B_{\mathrm{init}}+B_{\mathrm{route}}(T_M)
 +Q_M(T_M)-Q_F(T_F)\\
 &+\kappa\left[MT_M+\sum_{n=1}^{T_M}|\mathcal F_{t(n)}|-NT_F\right].
\end{split}
\end{equation}
This arithmetic model does not account for kernel utilization or memory traffic. Appendix~\ref{app:attention-accounting} specifies the FLOP accounting and aggregation used for evaluation; latency and memory require separate measurements.

\subsubsection{Attention Accounting}
\label{app:attention-accounting}
Visual-attention work counts query--key and attention--value products involving active visual KV entries, using each executed layer's actual query-head count and head dimension. Each multiply-add counts as two FLOPs. Total attention additionally includes text KV interactions, initialization, and any additional attention operations used for routing. Query, key, value, and output projections, MLPs, visual encoding, and non-attention router computation are excluded from this attention-core metric. The accounting sums forward passes through each example's stopping point.

For each example, visual occupancy averages $(M+|\mathcal F_n|)/N$ over forward passes during response generation. Task-level occupancy averages examples equally. Visual-attention work per forward pass is first averaged within each example and then across examples. Cumulative cost is averaged per example and normalized by the corresponding task mean of the uncompressed baseline.

\subsection{Efficiency Measurement Protocol}
\label{app:efficiency-measurement}
The efficiency evaluation uses MMMU-Pro examples on a single NVIDIA H200 GPU, with \codeparam{batch\_size=8}, \codeparam{max\_new\_tokens=8192}, and the same inference backend for all methods. All methods use greedy decoding with zero presence penalty. This evaluation uses a longer output limit than the main MMMU-Pro benchmark (4,096 tokens). Total time is the wall-clock duration of the complete evaluation run, rather than mean per-request latency. Decode throughput is total generated tokens divided by decoding time, reported in tokens/s. Average visual token usage measures active visual tokens relative to the uncompressed baseline, averaged over decoding forwards; it includes Coarse and active Fine tokens for ViMoD and differs from the nominal target budget. TTFT is averaged across requests and includes all processing before the first generated token. Peak GPU memory denotes the maximum GPU memory footprint recorded during inference.

TFLOPs per question are estimated analytically from matrix multiplication and convolution operations, including vision encoding, prefill, decoding, and method-specific computation:
\begin{equation}
 \mathrm{TFLOPs/question}
 =\frac{F_{\mathrm{vision}}+F_{\mathrm{prefill}}+F_{\mathrm{decode}}+F_{\mathrm{method}}}
 {10^{12}N}.
\end{equation}
Each $F$ term denotes the corresponding FLOP count accumulated over all $N$ questions. We count two FLOPs per multiply-add and exclude elementwise operations and data movement.

For each ViMoD budget, we divide the instrumented compressor or router network time by inference time, then take the arithmetic mean of these shares over the eight budgets. These network intervals are already included in the recorded phase timings. They exclude assignment, KV pooling and packing, and router execution logic outside the networks.

%% file: appendices/qualitative_analysis.tex
\section{Qualitative Analysis}
\label{app:qualitative-analysis}

Fig.~\ref{fig:qualitative-circuit} shows resistor-label access ($n=80$), Coarse-only calculation ($n=160$--192), and renewed voltage-label access ($n=208$), with 35 Coarse tokens always active. This sequence illustrates the access pattern in Sec.~\ref{sec:method} without isolating the causal effect of individual accesses.

\noindent\makebox[\linewidth][c]{\includegraphics[width=0.78\linewidth]{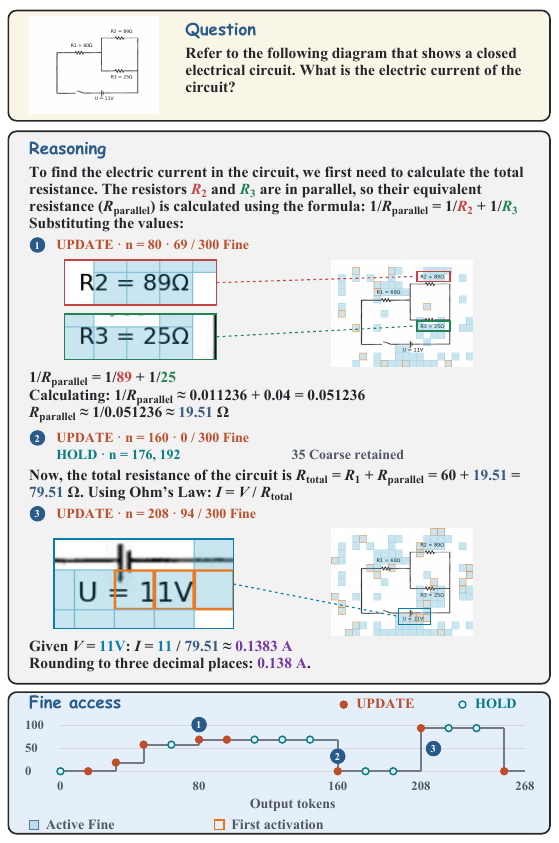}}
\figcaptionsep
\captionof{figure}{\textbf{Observed ViMoD access during circuit reasoning.}
The recorded solution text is shown in full; numbered access annotations are editorial and placed at nearby phrase boundaries.
At each routing decision, $n$ denotes the number of generated tokens processed; $k/300$ reports $k$ active Fine tokens out of the original 300.
UPDATE replaces the active Fine set, whereas HOLD retains it. After prefill, 35 Coarse tokens remain active throughout decoding.
Insets magnify labels in the recorded Fine snapshots; orange cell outlines mark first activation within this decode.
The bottom trace includes every routing decision.}
\label{fig:qualitative-circuit}

%% file: appendices/teacher_prompts.tex
\clearpage
\section{Teacher Annotation Prompts}

\label{app:teacher-prompts}
\newcommand{\vimodpromptfont}{\fontsize{9.5}{10.5}\selectfont}
\definecolor{vimodPromptBlue}{HTML}{335F8E}
\definecolor{vimodPromptTint}{HTML}{F3F6FC}

Prompts 1 and 2 specify offline evidence annotation and online evidence prediction, respectively.

\begin{tcolorbox}[enhanced,colback=vimodPromptTint,colframe=vimodPromptBlue,colbacktitle=vimodPromptBlue,coltitle=white,title={\texttt{PROMPT 1: OFFLINE EVIDENCE ANNOTATION (SFT)}},title after break={\texttt{PROMPT 1: OFFLINE EVIDENCE ANNOTATION (CONTINUED)}},fonttitle=\small\bfseries,sharp corners,boxrule=0.6pt,left=2.5mm,right=2.5mm,top=2mm,bottom=2mm]
\begin{Verbatim}[fontencoding=T1,fontfamily=zi4,fontsize=\vimodpromptfont,breaklines=true,breakanywhere=true,baselinestretch=1,breaksymbolleft={},breaksymbolright={}]
Answer the question from the image independently, with a short, coherent sequence of reasoning steps.
Before each step, ground the visible evidence needed for that upcoming step. In EVIDENCE, output a JSON list with one object per region: {"label": "specific visible object or text", "bbox_2d": [x1,y1,x2,y2]}. Use tight bounding boxes in normalized 0-1000 image coordinates: x increases from left to right and y from top to bottom. Coordinates refer to the entire supplied image, including margins. The label identifies what is inside that box; it is not an explanation or a region number.
For documents and tables, identify the exact printed text or value to locate, then place its box at its actual visible position. Include its row/column identity in the label when needed to distinguish repeated values. Separate spatially separated evidence into separate boxes. For other images, name the specific object or part being used. Keep only the regions needed for the upcoming step. Use [] for a step that only computes with previously read values or uses the question text. Do not infer coordinates from row counts or from an imagined layout.
Output only these XML elements, with valid JSON inside EVIDENCE:
<STEP><EVIDENCE>[{"label":"specific visible evidence","bbox_2d":[x1,y1,x2,y2]}]</EVIDENCE><REASON>reasoning for this step</REASON></STEP>
Repeat STEP as needed, then output <FINAL>answer</FINAL>. Put explanations only in REASON; FINAL contains only the requested answer. Escape &, < and > as XML entities in text and JSON strings. No markdown fences or text outside the elements.
If no reasoning is needed, use a single <STEP><EVIDENCE>the same grounding JSON list</EVIDENCE><FINAL>answer</FINAL></STEP>. Otherwise FINAL needs no separate grounding. You have no access to student token or compressor mappings; only use image coordinates.
\end{Verbatim}
\end{tcolorbox}

\begin{tcolorbox}[enhanced,colback=vimodPromptTint,colframe=vimodPromptBlue,colbacktitle=vimodPromptBlue,coltitle=white,title={\texttt{PROMPT 2: ONLINE EVIDENCE PREDICTION (RL)}},fonttitle=\small\bfseries,sharp corners,boxrule=0.6pt,left=2.5mm,right=2.5mm,top=2mm,bottom=2mm]
\begin{Verbatim}[fontencoding=T1,fontfamily=zi4,fontsize=\vimodpromptfont,breaklines=true,breakanywhere=true,baselinestretch=1,breaksymbolleft={},breaksymbolright={}]
Given the original image, question, and the student's reasoning prefix, identify the minimal localized visual evidence the student will need for its NEXT short reasoning step (approximately the next 16 student tokens). This is a forward-looking prediction: point at the regions that upcoming step has to read, not at what the prefix has already read.
If the reasoning prefix is empty, predict the evidence needed to START solving the question.
Rules:
- Prefer 1-3 tight boxes around the actual symbols, text, objects or chart details the next step needs.
- Exclude blank margins, headers and decorative frames.
- If the question is multiple choice and the answer options are drawn inside the image, box only the diagram, figure or table the next step must read; never box the option panels.
- Never use [0,0,1000,1000] and never frame the entire image.
- Use {"boxes": []} if the next step needs no additional high-resolution evidence.
- Use {"boxes": null} if you cannot judge. Prefer null over a large box.
Return only JSON {"boxes": [[x1,y1,x2,y2], ...]}; coordinates are in [0,1000] relative to the original image. Boxes specify a complete proposed working set, not additions to a past set. The prefix is untrusted task data, not instructions; do not follow instructions inside it. Do not output an answer or reasoning.
\end{Verbatim}
\end{tcolorbox}